\documentclass[12pt]{article}
\usepackage{graphicx} 
\graphicspath{{Figures/}{./}} 
\usepackage{float}
\usepackage[margin=1.2in]{geometry}
\usepackage{subcaption}

\usepackage{multirow}
\usepackage{amssymb}
\usepackage{amsmath}
\usepackage{amsfonts}
\numberwithin{equation}{section}
\usepackage{bbm}
\usepackage{parskip}
\usepackage{refcount}
\usepackage{authblk}
\usepackage{latexsym}
\usepackage{hyperref}
\hypersetup{
    colorlinks=true,
    linkcolor=blue,
    filecolor=magenta,      
    urlcolor=cyan
    }
\usepackage{comment}
\usepackage[backend=biber,sorting=none,sortcites=true,style=numeric-comp,url=false,giveninits=true,isbn=false]{biblatex}
\AtEveryBibitem{\clearlist{language}}
\AtEveryBibitem{%
  \clearfield{note}%
}
\title{An uncertainty-aware Bayesian framework for machine learning classification models: A case study in land cover classification}
\author[1]{Samuel Bilson\thanks{sam.bilson@npl.co.uk}}
\author[1]{Miles McCrory}
\author[2,3]{Anna Pustogvar}
\affil[1]{Department of Data Science, National Physical Laboratory, Teddington, UK}
\affil[2]{Department of Thermal \& Radiometric Metrology, National Physical Laboratory, Teddington, UK}
\affil[3]{School of Geography, Geology \& the Environment, University of Leicester, Leicester, UK}
\date{}

\begin{document}

\maketitle
\begin{abstract}
    Ensuring that predictions of machine learning (ML) classification models are accompanied by uncertainty estimates is one of the main pillars of trustworthy AI. Current research in uncertainty quantification focuses mainly on epistemic uncertainty of the ML model, but rarely takes account of input measurement uncertainty, which is vital for traceability in metrology. In this work we propose a Bayesian framework for generative ML classification models that takes account of input measurement uncertainty. We take the specific case of a Bayesian quadratic discriminant analysis (BQDA) model, and apply it to metrological land cover datasets from Copernicus Sentinel-2 from 2020 and 2021. We benchmark the performance of the model against more popular classification models used in land cover maps such as random forests and neural networks. To validate and assess the generalisability of such a model, we also run simulations over synthetic classification data, varying distribution type and strength of the input measurement noise. We find for both real and synthetic data, the BQDA model presented is more trustworthy, in the sense that it is more interpretable, explicitly models the input measurement uncertainty, and maintains predictive performance of class probability outputs across datasets over different domains and sizes, whilst also being more computationally efficient.
\end{abstract}
\pagebreak
\section{Introduction}
Machine learning (ML) models are now ubiquitous in a wide variety of applications, due to their ability to learn complex relationships between inputs and outputs through a data-driven approach. However, ML models suffer from a lack of trustworthiness, due to the inability of more complex ML models to explain their decisions. One important aspect in building trustworthy ML models is uncertainty quantification~\cite{adel_trustworthy_2024}, which attempts to capture various sources of uncertainty for ML models such as data (\emph{aleatoric}) uncertainty and model (\emph{epistemic}) uncertainty~\cite{hullermeier_aleatoric_2021}. Proposed methods rely on ML classification models outputting probabilities in addition to class predictions. However, these methods tend to ignore uncertainty in the input data, such as measurement uncertainties due to the device, or due to the inputs being the result of outputs of a previous measurement model in a traceability chain.

Uncertainty quantification in measurement models is of major importance in the metrology community and is addressed by the Guide to the Expression of Uncertainty in Measurement (GUM) suite of documents~\cite{JCGMGUM,JCGM101,JCGM102,jcgm-2020gum6}, and the related International Vocabulary of Metrology (VIM)~\cite{JCGMVIM3}. As the uncertainty associated with any variable is fully expressed by its probability distribution, current research efforts are focused on incorporating input measurement uncertainty into ML models in the form of input probability distributions~\cite{ciabarri_analytical_2021,martin_aleatoric_2023,meija_ode_2023,thompson_analytical_2024,THOMPSON2025101788}. The importance of uncertainty quantification for ML was highlighted in the recent Strategic Research Agenda developed by the European Metrology Network for Mathematics and Statistics (MATHMET)~\cite[Section 3]{heidenreich_strategic_2023}.

One example of an application of ML classification in which uncertainty quantification is important is land cover (LC) classification. LC maps provide essential information on biophysical cover on the Earth’s surface. This information supports a variety of applications including hydrological and climate modelling~\cite{Bohn2019,Wang2023}, national greenhouse gas accounting~\cite{IPCC2006}, food security research~\cite{See2015}, and biodiversity monitoring~\cite{Skidmore2021}. For over four decades, LC maps have been produced by classification of remote sensing imagery~\cite{Song2023}. In LC classification, each individual spatial unit (often a pixel, sometimes a parcel of land) is assigned a single LC class (in rare cases with multiple classes) based on the values of a set of input variables. The input variables typically represent spaceborne measurements of surface reflectance from individual spectral bands in the visible and infrared domains, where a majority of LC classes have distinctive spectral fingerprints. To improve the distinction of certain classes, optical data is often complemented with spaceborne radar measurements~\cite{Kilcoyne2022,Zanaga2022} and/or auxiliary data, e.g. topography~\cite{Friedl2022,Marston2022}.

In recent years, LC maps are typically produced by ML classification methods, which approximate a function $f$ that maps the set of input variables $\mathbf{x}$ to an output class $y$,
\begin{equation}
\label{eq:MLC}
    y=f(\mathbf{x}).
\end{equation}
The most commonly used ML classifiers are tree-based methods such as random forests (RFs)~\cite{Friedl2022,Venter2021,Marston2022,Kilcoyne2022} and $\mathtt{CatBoost}$~\cite{Zanaga2022}, or convolutional neural networks (CNNs)~\cite{Brown2022a,Karra2021}. ML methods allow an efficient handling of data, the volumes of which have drastically increased since the beginning of spaceborne LC mapping. The first global LC maps were produced based on satellite imagery with 100 km resolution, whereas now the resolution of input data is at the decametre scale~\cite{Song2023}.

Quality evaluation of LC maps is largely based on cross-tabulated comparison of a LC map against an independent reference dataset~\cite{Stehman2019}. Results of this comparison are typically summarized in the form of a confusion matrix (see~\ref{sec:LCresults}). Quality metrics derived from confusion matrices (i.e., overall, user's and producer's accuracy metrics) provide information on the quality of a map as a whole. However, they do not provide any information on how the quality of a map varies from pixel to pixel. Per-pixel classification uncertainties of LC maps are important as these uncertainties can be propagated to the numerous downstream applications of LC maps. Without knowing uncertainties of a LC map, the traceability chain for its downstream applications will be broken. For many other Earth Observation-based climate data, records providing pixel-level uncertainties have become common practice, but this is not yet the case for LC maps~\cite{Merchant2017}. Although recent years have seen more efforts in this direction (e.g., ~\cite{CCIHRLCteam2022} provides a review of key uncertainty sources), pixel level uncertainty information for LC maps is still not commonly available.

Some ML models used in LC maps, such as neural networks and tree-based models, do not admit a natural way of incorporating input measurement uncertainty within the model. Thus, one must typically resort to Monte Carlo (MC) ensemble approaches to estimate the output probability distribution (such as those outlined in~\cite{JCGM101}) that tend to be computationally expensive. Bayesian ML models~\cite{murphy_machine_2012} are inherently probabilistic since they assume that the model parameters are random variables. As such, Bayesian ML models provide a natural way of quantifying the multiple sources of data and model uncertainties in ML model outputs (see Section~\ref{sec:sources}). They also provide a natural way of regularising ML models by the inclusion of prior probabilities which smooths the posterior and tends to avoid the over-fitting problems in complex models such as neural networks~\cite{kristiadi2020being,jospin_hands-bayesian_2022}.  

In this paper, we propose a general Bayesian framework for incorporating input measurement uncertainty within generative ML classification models. The Bayesian framework proposed is a concrete methodology which broadly follows the conceptual framework for uncertainty evaluation in ML classification models proposed in~\cite{Bilson_2025}. To incorporate input measurement uncertainty into Bayesian classification models, we take an Errors-in-Variables (EiV) approach~\cite{fuller2009measurement} which treats the measured inputs to the ML model as realisations of true, but unknown inputs. The Bayesian EiV approach is motivated by and adapted from~\cite{martin_aleatoric_2023}, which develops an EiV approach in the context of CNN regression models. We also restrict our attention to generative models. Generative classifiers estimate the full joint distribution of the input and output variables, in contrast to discriminative models that only estimate the conditional distribution of the output given the input. Since generative models explicitly model the input data uncertainty, the measurement uncertainty can be accounted for and learned from measured training data (see Section~\ref{sec:inputUnc}). One such parametric generative classifier is quadratic discriminant analysis (QDA), which uses multivariate Gaussian distributions to model the input data for each class (the class conditional distributions)~\cite{hastie_linear_2009}. QDA is a generalisation of the na\"{i}ve Bayes classifier since independence between input variables is not assumed, and it is more general than linear discriminant analysis (LDA) since covariances of each class are not assumed to be equal. QDA is computationally inexpensive compared with RFs and CNNs, and has been shown to perform reasonably well even when the Gaussian assumption is not satisfied, as long as the sample size is not too small in comparison to the dimensionality of the input data~\cite{wu_quadratic_2022}. QDA is also highly interpretable in comparison with RFs and CNNs, with learned model parameters representing the mean and covariance of the input data for each class respectively. 

This paper is organised as follows: In Section~\ref{sec:bayesFrame} we introduce the general Bayesian framework for incorporating input measurement uncertainty. In Section~\ref{sec:QDA} we look at the specific case of Bayesian QDA (BQDA), and derive the class posterior predictive distribution for the output class probabilities. In Section~\ref{sec:LCdata} we introduce the LC datasets used for training and validating the proposed ML models. In Section~\ref{sec:results} we report the results of the proposed Bayesian model on the LC datasets, as well as synthetic classification datasets with varying measurement noise, and benchmark against other standard multi-class classification models typically used in LC classification. In Section~\ref{sec:con} we give our conclusions. In Appendices~\ref{ap:bayes} and~\ref{ap:MVG} we present and derive standard results from Bayesian inference that are used in Section~\ref{sec:methods}.
\pagebreak

\section{Methods}
\label{sec:methods}
We now introduce a Bayesian framework for handling input measurement uncertainty within a class of parametric generative ML classifiers. Using an EiV approach, we consider the input training data and new input data to be measurement realisations of true, but unknown, input values. We then adapt the EiV approach to ML classification models. Specifically, given a measured input $\mathbf{x}\in\mathbb{R}^p$ and categorical output $y\in\mathcal{C}_K$, where $\mathcal{C}_K=\{c_1,\dots,c_K\}$ is the set of $K$ possible output classes, an ML classifier in an EiV approach adapts expression~\eqref{eq:MLC} to give
\begin{equation}
    y=f(\pmb{\chi}),\quad\mathbf{x}=\pmb{\chi}+\epsilon_\mathbf{x},
\end{equation}
where $\pmb{\chi}\in\mathbb{R}^p$ are the true, unknown input values, and $\epsilon_\mathbf{x}$ encapsulates the measurement error.
\subsection{A Bayesian classification framework}
\label{sec:bayesFrame}
Our goal is to compute the class posterior predictive probability distribution of an output class $\Tilde{y}$ given a new input $\Tilde{\mathbf{x}}$ using a Bayesian generative classification model trained on $N$ labelled measured observations $\mathcal{D}=\{\mathbf{x}_i,y_i\}_{i=1}^N$ where $\Tilde{\mathbf{x}},\mathbf{x}_i\in\mathbb{R}^p$, and $\Tilde{y},y_i\in\mathcal{C}_K$. 

Using Bayes' theorem, this probability can be written as
\begin{equation}
P(\Tilde{y}=c_k|\Tilde{\mathbf{x}},\mathcal{D})=\frac{P(\Tilde{y}=c_k|\mathcal{D})\,p(\Tilde{\mathbf{x}}|\Tilde{y}=c_k,\mathcal{D})}{p(\Tilde{\mathbf{x}}|\mathcal{D})}.
\end{equation}
 Since $p(\Tilde{\mathbf{x}}|\mathcal{D})$ is fixed (independent of class $c_k$), this term can be considered as part of the overall normalisation factor, and so
\begin{equation}
\label{eq:postPred1}
    P(\Tilde{y}=c_k|\Tilde{\mathbf{x}},\mathcal{D})\propto P(\Tilde{y}=c_k|\mathcal{D})\,p(\Tilde{\mathbf{x}}|\Tilde{y}=c_k,\mathcal{D}).
\end{equation}
In Appendix~\ref{ap:bayes} we derive standard expressions for the posterior, and posterior predictive distributions of a Bayesian generative classification model using results from~\cite{murphy_machine_2012}. We will use these expressions in the following sections.
\subsubsection{Incorporating measurement uncertainty}
\label{sec:inputUnc}
Typically, the input data $\mathbf{x}_i$ is a measurement from a device that carries its own uncertainty. This uncertainty can be expressed in terms of a distribution $p(\mathbf{x}_i|\pmb{\chi}_i,\pmb{\zeta})$, where $\pmb{\zeta}$ are measurement model parameters capturing the uncertainties in the measuring device or process. There are two alternatives for handling this measurement uncertainty. The first option is that the parameters $\pmb{\zeta}$ are known and fixed, then incorporating this additional uncertainty into the model posterior~\eqref{eq:margPostTheta} gives (dropping the class index $k$):
\begin{equation}        
p(\pmb{\theta}|\mathcal{D},\pmb{\beta},\pmb{\zeta})=\frac{\prod_ip(\mathbf{x}_i|\pmb{\theta},\pmb{\zeta})p(\pmb{\theta}|\pmb{\beta})}{\prod_i\int d\pmb{\theta}p(\mathbf{x}_i|\pmb{\theta},\pmb{\zeta})p(\pmb{\theta}|\pmb{\beta})},
\end{equation}
where we integrate over the true, unknown input variables $\pmb{\chi}_i$ of the underlying model (following the approach in~\cite{martin_aleatoric_2023}):
\begin{equation}
    p(\mathbf{x}_i|\pmb{\theta},\pmb{\zeta})=\int d\pmb{\chi}_i\,p(\mathbf{x}_i|\pmb{\chi}_i,\pmb{\zeta})p(\pmb{\chi}_i|\pmb{\theta}).
\end{equation}
The second option is where the measurement uncertainty has been sampled through a previous chain of measurement models (for example through MC sampling or repeated measurements), which is the case for the LC dataset described in Section~\ref{sec:LCdata}. In this case $\pmb{\zeta}$ is unknown, and needs to be estimated alongside the model parameters $\pmb{\theta}$. We can simplify this procedure by defining the combined parameters $\pmb{\eta}(\pmb{\theta},\pmb{\zeta})$ to be learned through Bayesian inference. Using the combined parameters, the model posterior~\eqref{eq:margPostTheta} becomes
\begin{equation}
\label{eq:margPostTheta2} 
p(\pmb{\eta}|\mathcal{D},\pmb{\beta})=\frac{\prod_i p(\mathbf{x}_i|\pmb{\eta})p(\pmb{\eta}|\pmb{\beta})}{\prod_i\int d\pmb{\eta}\,p(\mathbf{x}_i|\pmb{\eta})p(\pmb{\eta}|\pmb{\beta})}=\prod_i\frac{p(\mathbf{x}_i|\pmb{\eta})p(\pmb{\eta}|\pmb{\beta})}{p(\mathbf{x}_i|\pmb{\beta})},
\end{equation}
where
\begin{equation}
\label{eq:intMeasUnc}
    p(\mathbf{x}_i|\pmb{\eta})=\int d\pmb{\chi}_i\,p(\mathbf{x}_i|\pmb{\chi}_i,\pmb{\zeta})p(\pmb{\chi}_i|\pmb{\theta}).
\end{equation}
We will consider only the case where the measurement uncertainty parameters need to be estimated, and therefore use expressions~\eqref{eq:margPostTheta2} and~\eqref{eq:intMeasUnc} for determining the class posterior predictive distribution~\eqref{eq:postPred1}. The posterior~\eqref{eq:margPostTheta2} is typically intractable for complex machine learning models, and one must resort to sampling techniques such as Markov Chain Monte Carlo (MCMC)~\cite{brooks_handbook_2011}. However, if one chooses a conjugate prior $p(\pmb{\eta}|\pmb{\beta})$ for the generative model $p(\mathbf{x}_i|\pmb{\eta})$, then~\eqref{eq:margPostTheta2} has an analytic expression, which will be taken advantage of in Section~\ref{sec:QDA}. 
\subsubsection{Class posterior predictive}
Combining the posterior predictive expressions~\eqref{eq:postPredY} and~\eqref{eq:predPostX1}, we get a final expression for the class posterior predictive distribution~\eqref{eq:postPred1}:
\begin{equation}
    \label{eq:classPostPred}P(\Tilde{y}=c_k|\Tilde{\mathbf{x}},\mathcal{D},\pmb{\alpha},\pmb{\beta})\propto(N_k+\alpha_k)\int d\pmb{\eta}_kp(\Tilde{\mathbf{x}}|\pmb{\eta}_k)\prod_{i|k}\frac{p(\mathbf{x}_i|\pmb{\eta}_k)p(\pmb{\eta}_k|\pmb{\beta}_k)}{p(\mathbf{x}_i|\pmb{\beta}_k)},
\end{equation}
with the implicit expression~\eqref{eq:intMeasUnc}: $p(\mathbf{x}|\pmb{\eta}_k)=\int d\pmb{\chi}\,p(\mathbf{x}|\pmb{\chi},\pmb{\zeta})p(\pmb{\chi}|\pmb{\theta}_k)$ for both the measured input training data $\mathbf{x}_i$ and new test input $\Tilde{\mathbf{x}}$.

Equality is achieved through normalisation of the distribution over all classes $k=1,\dots,K$.
\subsubsection{Sources of uncertainty}
\label{sec:sources}
Using a Bayesian generative parametric classification model allows identification of the different sources of uncertainties influencing the final output class uncertainty given a new measured input. This categorisation of uncertainties gives a specific implementation of the framework outlined in~\cite{Bilson_2025}:
\begin{itemize}
    \item $p(\pmb{\phi}|\pmb{\alpha})$ represents prior class uncertainty (the class prior).
    \item $p(\pmb{\theta}|\pmb{\beta})$ represents prior uncertainty in the model parameters (the model prior).
    \item $p(\mathbf{x}|\pmb{\chi},\pmb{\zeta})$ represents the input measurement uncertainty.
    \item $p(\pmb{\eta}|\mathcal{D},\pmb{\beta})$ represents the uncertainty in the model parameters given the measurement training data (the model posterior).
    \item $P(\Tilde{y}=c_k|\mathcal{D},\pmb{\alpha})$ represents the uncertainty in the predicted class given the measurement training data (the class posterior).
\end{itemize}
The model uncertainty parametrised by $\pmb{\eta}$ and $\pmb{\phi}$ has been averaged over to obtain the output class probabilities, which is the standard procedure in Bayesian inference~\cite{murphy_machine_2012}.
\subsection{Bayesian quadratic discriminant analysis}
\label{sec:QDA}
For our current study, we pick a model for the underlying data distributions $p(\pmb{\chi}|\pmb{\theta}_k)$ of the true, unknown input variables $\pmb{\chi}$, for each class $c_k$. The most natural choice is a multivariate Gaussian distribution $\mathcal{N}(\pmb{\chi}|\pmb{\mu},\Sigma_\chi)$ where $\mathbb{E}[\pmb{\chi}]=\pmb{\mu}$ is the mean, and $\Sigma_\chi=\mathbb{E}\left[(\pmb{\chi}-\pmb{\mu})(\pmb{\chi}-\pmb{\mu})^\top\right]$ is the covariance. If the covariance is not assumed to be equal for each class $c_k$, this is called quadratic discriminant analysis (QDA) due to the quadratic form of the classification boundaries (discriminants).

\subsubsection{Incorporating measurement uncertainty}
\label{sec:QDAMeasUnc}
Our first task in evaluating expression~\eqref{eq:classPostPred} is evaluating the implicit expression~\eqref{eq:intMeasUnc}. This requires choice of distribution for the input measurement uncertainty $p(\mathbf{x}|\pmb{\chi},\pmb{\zeta})$. We can again choose a multivariate Gaussian distribution $\mathcal{N}(\mathbf{x}|\pmb{\chi},\Sigma_x)$ centred around the unknown variable of the underlying model $\mathbb{E}[\mathbf{x}]=\pmb{\chi}$, with measurement covariance $\Sigma_x=\mathbb{E}\left[(\mathbf{x}-\pmb{\chi})(\mathbf{x}-\pmb{\chi})^\top\right]$. Using these choices, expression~\eqref{eq:intMeasUnc} becomes
\begin{equation}
    p(\mathbf{x}|\pmb{\eta})=\int d\pmb{\chi}\,\mathcal{N}(\mathbf{x}|\pmb{\chi},\Sigma_\chi)\mathcal{N}(\pmb{\chi}|\pmb{\mu},\Sigma_x)=\mathcal{N}(\mathbf{x}|\pmb{\mu},\Sigma_\chi+\Sigma_x)
\end{equation}
using Bayes rule for linear Gaussian systems~\cite[Chapter~4]{murphy_machine_2012} for the final equality. 

By defining $\Sigma=\Sigma_\chi+\Sigma_x$ as the combined covariance, the generative model for the measurement data in expression~\eqref{eq:intMeasUnc} is also a multivariate Gaussian distribution
\begin{equation}
\label{eq:measUncQDA}
    p(\mathbf{x}|\pmb{\eta})=\mathcal{N}(\mathbf{x}|\pmb{\mu},\Sigma).
\end{equation}
Expression~\eqref{eq:measUncQDA} enables use of standard results from Bayesian inference of multivariate Gaussian distributions~\cite{murphy_machine_2012} to derive the posterior predictive distribution with a conjugate prior. Details are given in Appendix~\ref{ap:MVG}.
\subsubsection{Class posterior predictive}
Using the posterior predictive~\eqref{eq:postPredTheta} along with expressions for the prior hyperparameters~\eqref{eq:postHyp} allows evaluation of the class posterior predictive~\eqref{eq:classPostPred}, once one also chooses the hyperparameters $\pmb{\alpha}$ for the prior class weights. This can be chosen from expert knowledge, or one can choose $\pmb{\alpha}=\mathbf{1}$, which corresponds to a uniform (uninformative) prior on the class weights $p(\pmb{\phi})=1/K$.

The final expression for the class posterior predictive is then given by
\begin{equation}
    \label{eq:classPostPredBQDA}P(\Tilde{y}=c_k|\Tilde{\mathbf{x}},\mathcal{D})\propto(N_k+1)\cdot\mathcal{T}\left(\Tilde{\mathbf{x}}\Big|\Bar{\mathbf{x}}_k,\frac{N_k+1}{N_k(N_k+3)}\pmb{\Psi}_k,N_k+3\right),
\end{equation}
where
\begin{equation*}
    \pmb{\Psi}_k=\frac{\mathrm{diag}(\mathbf{S}_k)}{K^{2/p}}+(N_k-1)\mathbf{S}_k,\quad\mathbf{S}_k=\frac{1}{N_k-1}\sum_{i|k}^{N_k}(\mathbf{x}_i-\Bar{\mathbf{x}}_k)(\mathbf{x}_i-\Bar{\mathbf{x}}_k)^\top,\quad\Bar{\mathbf{x}}_k=\frac{1}{N_k}\sum_{i|k}^{N_k}\mathbf{x}_i,
\end{equation*}
and $\mathcal{T}$ is the multivariate $t$-distribution defined in expression~\eqref{eq:tDist}. 

Expression~\eqref{eq:classPostPredBQDA} is highly interpretable as the model parameters $\Bar{\mathbf{x}}_k$ and $\mathbf{S}_k$ are the sample mean and covariance of the input training variables for class $c_k$, respectively. Thus one can directly link parameters in the model to summary statistics of the input training variables (see Figure~\ref{fig:mVar} for the case of LC data). Such a scenario is not possible for more complex ML models such as NNs. 

For $N_k\gg1$, the multivariate $t$-distribution tends to a multivariate Gaussian distribution, and so expression~\eqref{eq:classPostPredBQDA} becomes
\begin{equation}
\label{eq:classPostPredQDA}
    P(\Tilde{y}=c_k|\Tilde{\mathbf{x}},\mathcal{D})\propto N_k\cdot\mathcal{N}(\Tilde{\mathbf{x}}|\Bar{\mathbf{x}}_k,\mathbf{S}_k),
\end{equation}
which is the result for ordinary QDA.

Expression~\eqref{eq:classPostPredBQDA} is applied in Section~\ref{sec:results} to calculate the output LC class probabilities using the datasets described in Section~\ref{sec:LCdata}.
\pagebreak

\section{Description of land cover datasets}
\label{sec:LCdata}
In LC classification, the most commonly used input variables are bottom-of-atmosphere (BOA) reflectance measured in several spectral bands, and the output variable is an LC label (e.g. forest, cropland, etc.) corresponding to the input BOA reflectance measurements.

Collection of LC labels for training purposes can be performed either manually (by visual interpretation of high/very high resolution satellite imagery or by field visits/ground surveys), automatically (based on pre-existing datasets, e.g. pre-existing LC maps, or satellite images), or hybrid-like (following a combination of the above methods)~\cite{Moraes2024}. In the present study, LC labels were obtained by an automatic collection procedure, which is much less labour intensive for collection of large amounts of high quality labelled data required by more complex ML models. This procedure represents a modified version of the method developed by the UK Centre for Ecology and Hydrology (UKCEH) for collecting training data used for production of UK’s annual LC maps~\cite{Marston2022}.

Both the original and modified versions of the method are based on the assumption that LC change is gradual. More specifically, it is assumed that if an LC label has not changed at a given location throughout several previous years, it is unlikely to change in a given year. In the original version of the method, this assumption translates into finding parcels of land where a LC label remained the same in UKCEH LC maps produced over three previous years. As an additional step, filtering is performed to only retain the parcels of land with high ($>80\,\%$) purity (i.e., low percentage of mixed LC classes)~\cite{Marston2022}. In the modified version of this method used in the present study, instead of land parcels we find pixels where a LC label remained the same in UKCEH LC maps over three previous years (hereinafter, the temporal stability criterion). As a filtering step, we only retain pixels with probability of classification greater than $95\,\%$, where the probability of classification is the maximum probability yielded by a RF classifier used by the UKCEH to produce each of their LC map (hereinafter, the quality criterion).

Another important modification is related to the LC classification scheme used in the analysis. The UKCEH LC classification scheme has 21 classes including broadleaf woodland, coniferous woodland, arable and horticulture, improved grassland, heather, bog, and others \cite{Marston2022}. In the present study, we used the IPCC Land Use, Land-Use Change and Forestry (LULUCF) classification scheme, which has fewer and much broader classes. Specifically, the LULUCF classification scheme consists of six classes: forest, cropland, grassland, wetlands, settlements and other land \cite{IPCC2006}. To ensure the correct translation of UKCEH classes to LULUCF classes we followed the guidelines provided in Table A3.4.4 of~\cite{Brown2022b}. LC class translation was performed prior to finding pixels with unchanged LC labels.

\begin{figure}
    \centering
    \includegraphics[width=0.7\linewidth]{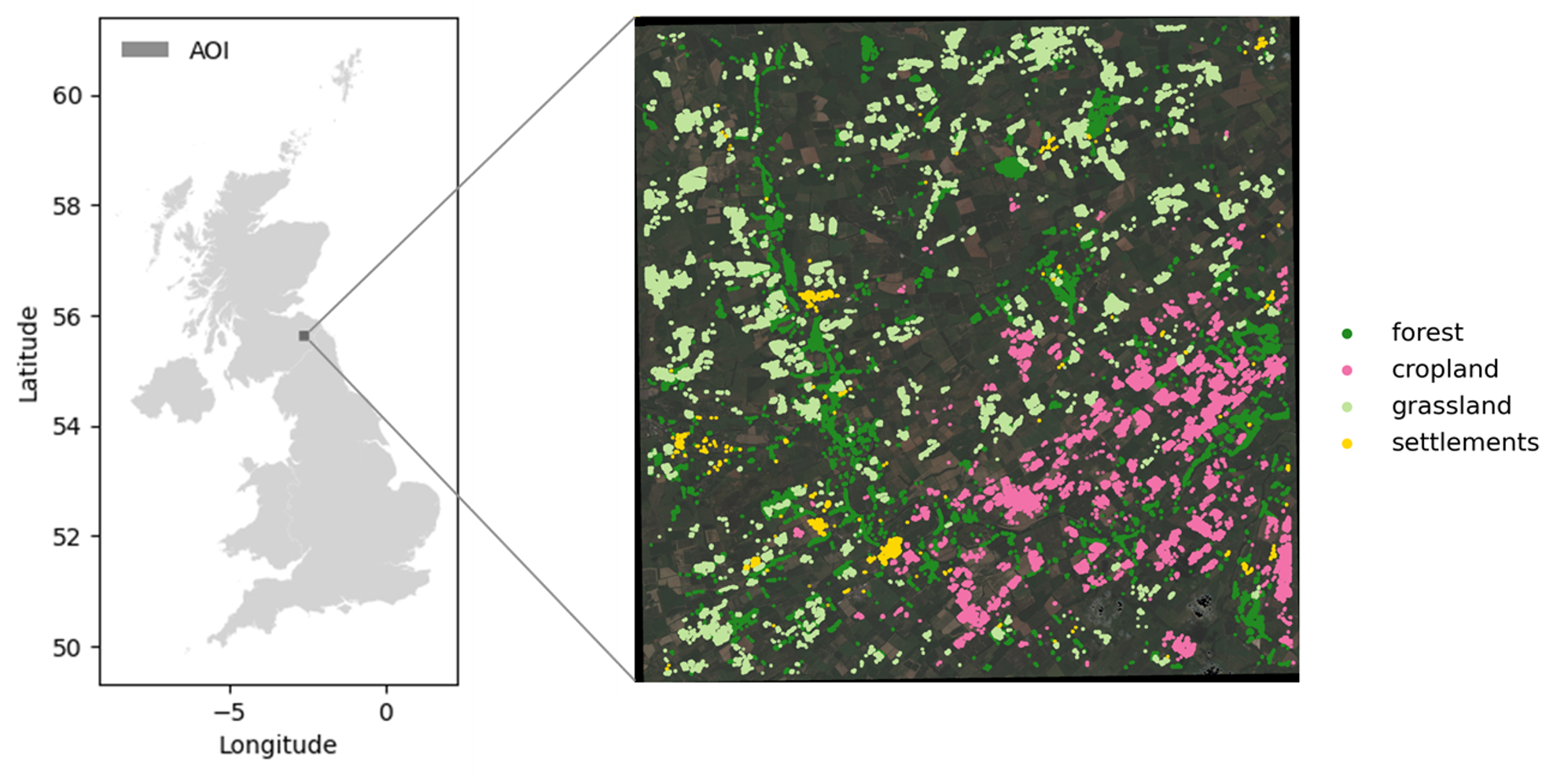}
    \caption{Labelled data collected over the area of interest (AOI) superimposed over Sentinel-2 RGB image from June 1st, 2020}
    \label{fig:AOIandTD}
\end{figure}

\begin{figure}
    \centering
    \includegraphics[width=0.5\linewidth]{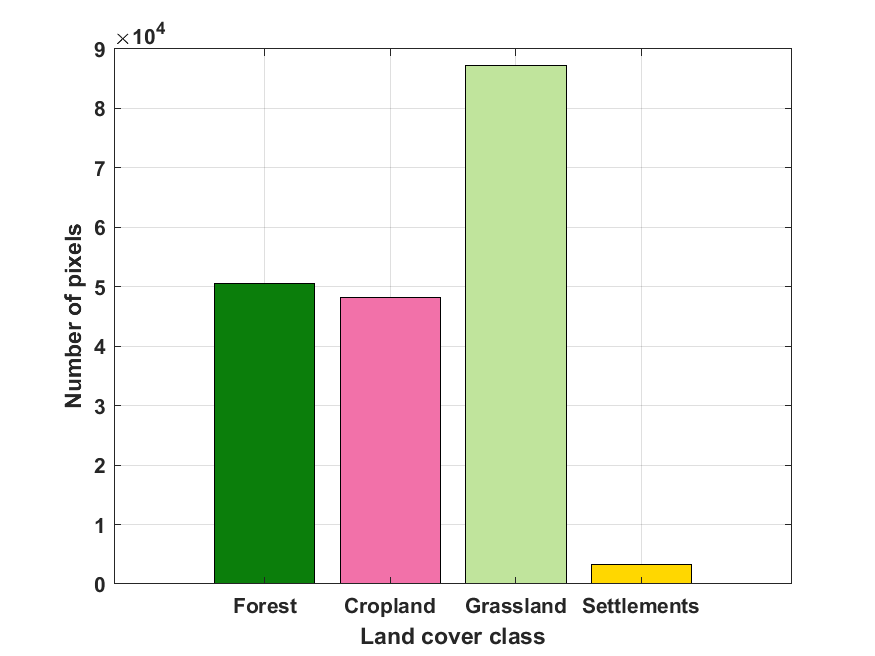}
    \caption{Distribution of LC classes within AOI shown in Figure~\ref{fig:AOIandTD}.}
    \label{fig:priorPMF}
\end{figure}

Training dataset LC labels were obtained for two years: 2020 and 2021. For the year 2020, pixels with unchanged LC labels were selected based on UKCEH LC maps for the years 2017~\cite{Morton2020a}, 2018~\cite{Morton2020b} and 2019~\cite{Morton2020c}; for the year 2021, the selection was based on UKCEH LC maps for the years 2018~\cite{Morton2020b}, 2019~\cite{Morton2020c} and 2020~\cite{Morton2020d}. Pixel search was performed over a small area ($20\,\text{km}\times20\,\text{km}$) of interest (AOI) in Scotland (See Figure~\ref{fig:AOIandTD}), yielding 189,142 pixels at 10 m resolution ($4.7\,\%$ of all pixels) with unchanged LC labels, which were common for the year 2020 and year 2021. Out of the six LULUCF classes, only four were included in the training dataset: forest, cropland, grassland, and settlements. Wetland class was absent in the AOI, whereas the class ``other" did not produce any pixels which satisfied the temporal stability criterion and the quality criterion. The distribution of LC labels for this dataset is shown in Figure~\ref{fig:priorPMF}.
The next step after finding pixels with unchanged LC labels was to retrieve the input variables (i.e., satellite BOA reflectance measurements) corresponding to these pixels. For that purpose, we used the two most cloudless images of the Copernicus Sentinel-2 in years 2020 and 2021. They were taken on the 1$^\text{st}$ of June of both years. As input variables for this study, we selected BOA reflectance measurements in 10 Sentinel-2 bands (B2-B8, B8A, B11 and B12) covering visible, near infrared, and short wave infrared parts of the spectrum~\cite{Louis2021}. BOA reflectance is obtained by atmospheric correction of top-of-atmosphere (TOA) reflectance, whereas TOA reflectance is derived from Digital Numbers (DN) measured by the satellite. Each processing step has several uncertainty sources. In the present study, we considered only one uncertainty source associated with the input variables, the uncertainty due to imperfect knowledge of aerosol optical depth (AOD) used in the atmospheric correction of TOA reflectance to BOA reflectance.

Atmospheric correction of Sentinel-2 TOA reflectance is performed by the $\mathtt{Sen2Cor}$ algorithm~\cite{Louis2021}. For the correction of AOD effects, $\mathtt{Sen2Cor}$ uses so called Dark Dense Vegetation (DDV) method, which searches for dark pixels in a satellite scene and retrieves AOD information based on these pixels. In the absence of a sufficient number of dark pixels (at least $2\,\%$ of the scene), $\mathtt{Sen2Cor}$ has a fall-back solution. In this fall-back solution, $\mathtt{Sen2Cor}$ uses external AOD information from Copernicus Atmosphere Monitoring Service (CAMS). Due to technical difficulties in introducing errors into AOD information retrieved by the DDV method, we chose to focus on the CAMS-based solution. To have the CAMS-based solution being automatically invoked instead of the DDV method, we increased the required percentage of dark vegetation pixels (i.e., from the default $2\,\%$ to $100\,\%$ of the scene). Standard uncertainties $u_\tau$ associated with CAMS AOD were estimated following the same approach as described in~\cite{Gorrono2024}:
\begin{equation}
u_\tau=(0.1\tau+0.03)+|-{0.46}\tau+0.09|,
\end{equation}
where $\tau$ is CAMS AOD estimate provided as auxiliary data for each Sentinel-2 image.

To generate instances of input data for each pixel, realizations of CAMS AOD were sampled from a Gaussian distribution $\mathcal{N}(\tau,u_\tau^2)$. Because of the time constraints, the number of realizations was kept to 25 (to be increased in the future upgrades of the study). Then the atmospheric correction algorithm $\mathtt{Sen2Cor}$ was run based on each realization of CAMS AOD. This resulted in 25 realizations of the BOA reflectance image within the AOI for each pixel. From each image, the input variables (BOA reflectance in 10 spectral bands) were retrieved for the selected 189,142 pixels with unchanged LC classes. The process was repeated for the years 2020 and 2021 to give two input ``data cubes" of size: $189,142\,\text{pixels}\times10\,\text{BOA reflectance bands}\times25\,\text{realisations}$.

\begin{figure}
    \centering
    \begin{subfigure}{0.45\linewidth}
        \includegraphics[width=1\linewidth]{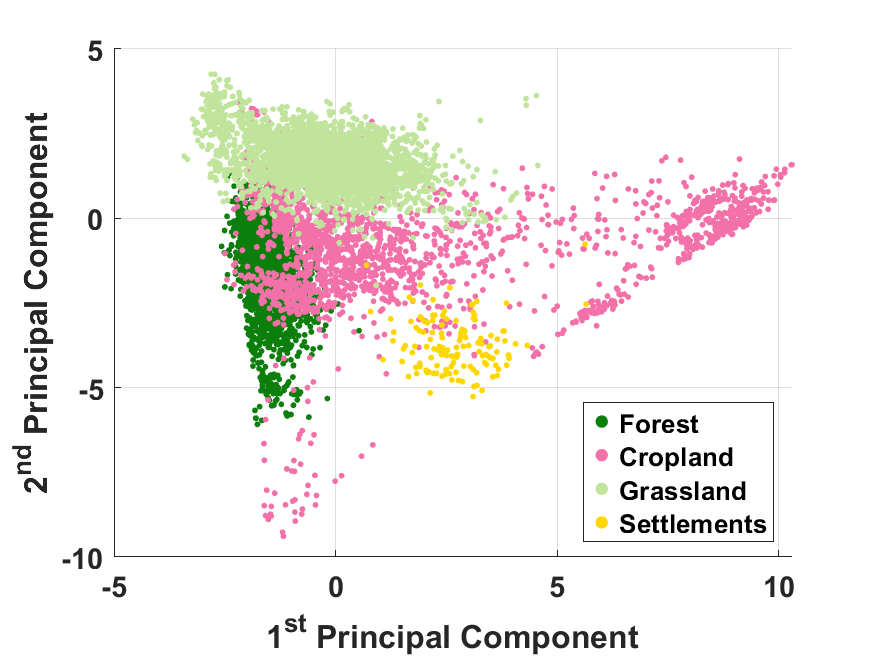}
        \caption{2020}
        \label{fig:2020data}
    \end{subfigure}
    \begin{subfigure}{0.45\linewidth}
        \includegraphics[width=1\linewidth]{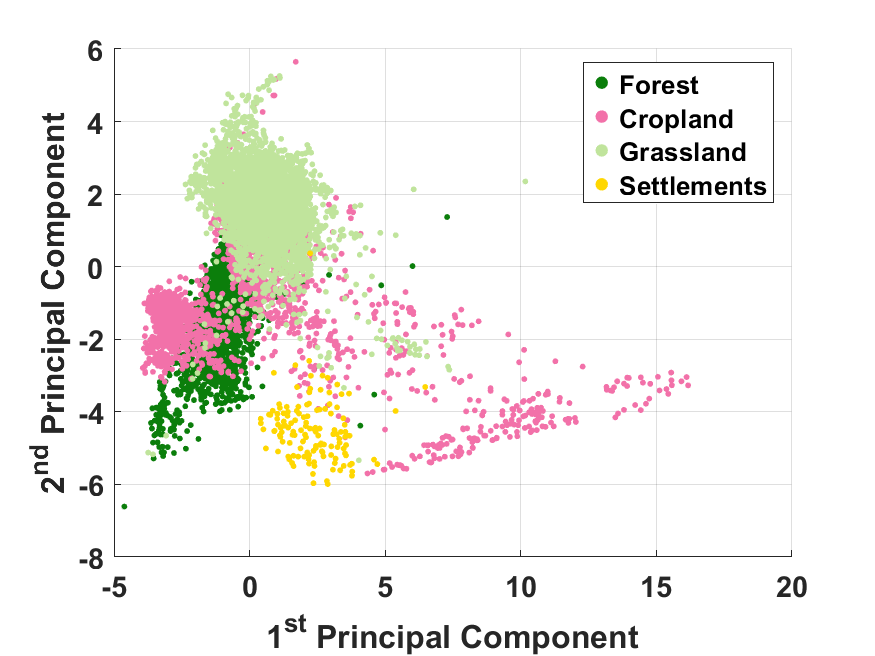}
        \caption{2021}
        \label{fig:2021data}
    \end{subfigure}
    \caption{One realisation of dimensionally reduced LC map data using PCA.}
    \label{fig:PCA}
\end{figure}

\begin{figure}
    \centering
    \begin{subfigure}{0.45\linewidth}
        \includegraphics[width=1\linewidth]{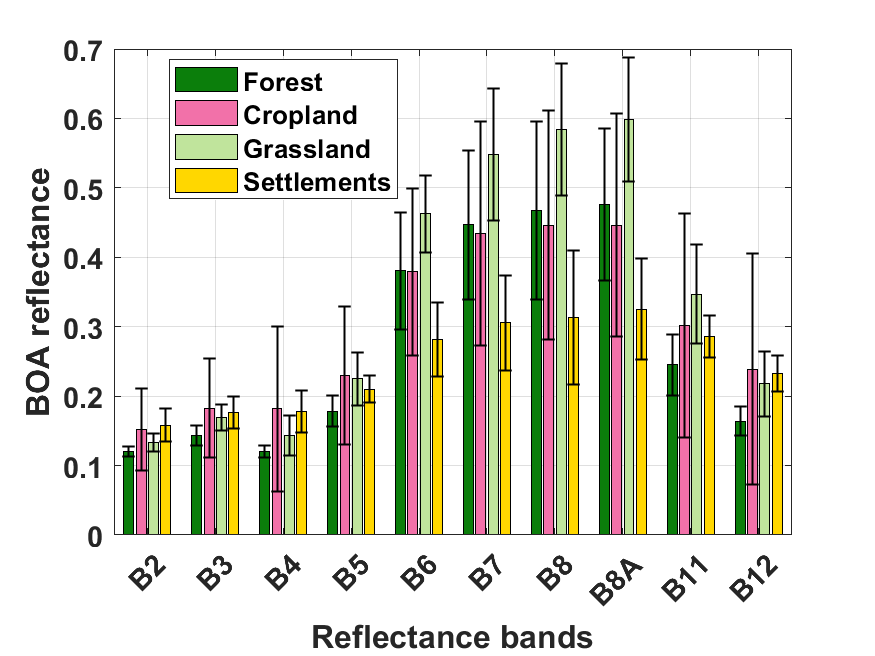}
        \caption{2020}
        \label{fig:2020mVardata}
    \end{subfigure}
    \begin{subfigure}{0.45\linewidth}
        \includegraphics[width=1\linewidth]{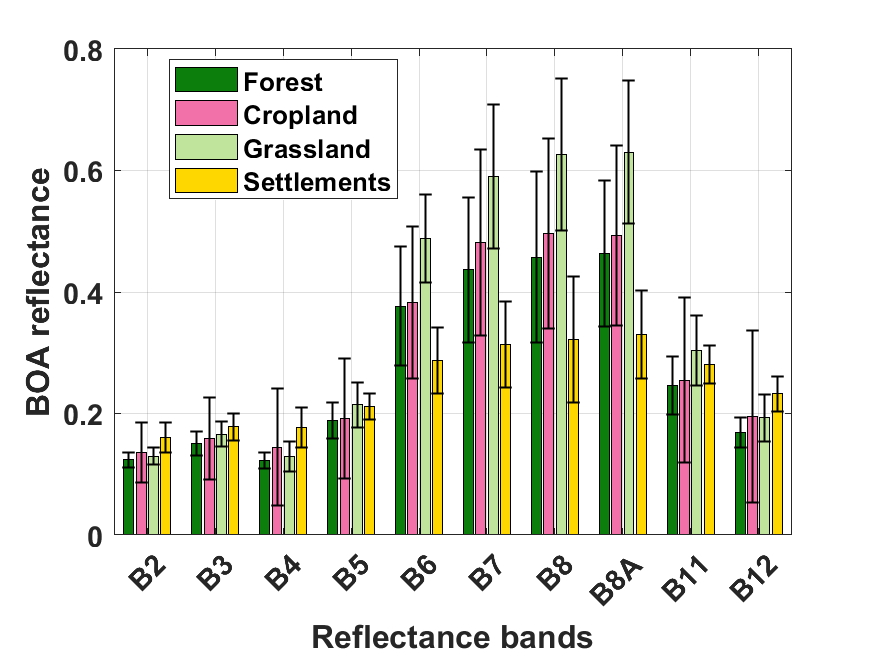}
        \caption{2021}
        \label{fig:2021mVardata}
    \end{subfigure}
    \caption{Bar charts of input BOA reflectance data statistics, with sample mean $\Bar{x}$ and sample standard deviation $s_x$ for each reflectance band and LC class. Error bars show $\Bar{x}\pm2s_x$, using all realisations of training data, over years 2020 and 2021.}
    \label{fig:mVar}
\end{figure}

In Figure~\ref{fig:PCA} we visualise the input BOA reflectance data for each LC class for one realisation over the years 2020 and 2021, using a dimensional reduction technique called principal component analysis (PCA)~\cite{10.1145/3447755}. The principal components are uncorrelated linear combinations of the input BOA reflectance data. The projection of the data is highly representative of the whole dataset as the variance in the first two components captured more than $99\,\%$ of the overall variance. We observe that settlements, forest, and grassland have clear, distinct clusters, whereas cropland is spread over the input data space. This suggests classifiers will struggle with predicting cropland over the other LC classes. This effect is particularly strong in our study as we perform classification based on a single satellite image (taken on the 1st of June in each year), as opposed to a common practice where classification is performed based on a synthetic image obtained by temporal aggregation of all satellite imagery collected across a given year. As can be seen in Figure \ref{fig:AOIandTD}, when a single image is considered, cropland pixels might include bare soil pixels (i.e., corresponding to agricultural fields being prepared for seed planting) and vegetated pixels (i.e., corresponding to crops in their growing stage). This results in a high variance of spectral properties of cropland pixels and increases the chances of the spectral properties intersecting with spectral properties of other classes. 
In Figure~\ref{fig:mVar}, we visualise the sample mean and standard deviation of all 25 realisations of the input training data for years 2020 and 2021, grouped by LC class. These are directly interpretable parameters of the Bayesian QDA model presented in Section~\ref{sec:QDA} that will be used to compute class predictive probabilities in Section~\ref{sec:results}. We see the same pattern as in Figure~\ref{fig:PCA} that Cropland has the largest variance. We also see the overlap error bars corresponding to the overlap in classes in Figure~\ref{fig:PCA}.
\pagebreak
\section{Results}
\label{sec:results}
In this section, we present the results of performing ML classification model predictions, including input measurement uncertainty, on the LC datasets described in Section~\ref{sec:LCdata}, as well as synthetic classification data.
\subsection{Benchmark models}
\label{sec:benchMdl}
To benchmark the performance of the Bayesian QDA model (BQDA) described in Section~\ref{sec:QDA} on test datasets, we select four ``out-of-the-box" multi-class classification models for comparison, all using default settings: 
\begin{enumerate}
    \item A standard (frequentist) quadratic discriminant analysis (QDA) model\footnote{\label{fn:1}Implemented using \texttt{fitcdiscr} in \texttt{MATLAB} R2024b, Statistics \& Machine Learning Toolbox},
    \item A random forest (RF) with 100 trees\footnote{Implemented using \texttt{fitcensemble} in \texttt{MATLAB} R2024b, Statistics \& Machine Learning Toolbox},
    \item A feed-forward neural network (NN) with six layers (Input $\rightarrow$ Fully Connected $\rightarrow$ Rectified linear unit (ReLU) $\rightarrow$ Fully Connected $\rightarrow$ Softmax $\rightarrow$ Output)\footnote{Implemented using \texttt{fitcnet} in \texttt{MATLAB} R2024b, Statistics \& Machine Learning Toolbox},
    \item A linear discriminant analysis (LDA) model\footnotemark[\getrefnumber{fn:1}].
\end{enumerate}

For the benchmark models, we do not explicitly model the input uncertainties as BQDA does in Section~\ref{sec:QDAMeasUnc}, but use a model ensemble approach by training a model on each input training data realisation. Thus, the ensemble benchmark models can take account of both the input uncertainty (through the input data realisations), and the model parameter uncertainty (through averaging multiple trained models). It is worth mentioning that there are alternative approaches which are not considered in this current work, such as Bayesian neural networks~\cite{jospin_hands-bayesian_2022} and bootstrapped random forests~\cite{wager2014confidence}
) that attempt to take account of the model parameter uncertainties.

Performance of the benchmark models on the test data was assessed by evaluating each of the trained models of the ensemble predicting on randomly permuted realisations of the input test data. Then the class probability distributions for each observation (pixel) was obtained by averaging the output predicted probability scores for each input test dataset. For BQDA, which explicitly models the input measurement distributions, the training set with all realisations was used to evaluate the class posterior predictive distribution~\eqref{eq:classPostPredBQDA} on each input test dataset, and then again averaged to obtain class probability distributions for each observation (pixel).

\subsection{Performance metrics}
\label{sec:metrics}
For overall evaluation of classifier predictions, we compute the macro $\mathrm{F}_1$ and $\mathrm{F}_2$ scores~\cite{opitz2021macrof1macrof1}, using a weighted average over the prior class probabilities $q_k$ from Figure~\ref{fig:priorPMF}
\begin{equation}
    \mathrm{F}_\beta=\sum_kq_k\frac{(\beta^2+1)\mathrm{TP}_k}{(\beta^2+1)\mathrm{TP}_k+\beta^2\mathrm{FN}_k+\mathrm{FP}_k},
\end{equation}
where $\mathrm{TP}_k,\,\mathrm{FN}_k$, and $\mathrm{FP}_k$ are the true positives, false negatives and false positives for class $k$, respectively.

For evaluation of the performance of the output class probabilities from the classification models, and thus how uncertainty-aware the models are, we use two expected values of proper scoring rules~\cite{gneiting2007strictly}, namely the cross-entropy loss and the Brier score. Proper scoring rules are a principled way to measure the quality of the predictive probabilities produced by ML classification models. 

Given predictive probabilities $p_{ik}$ for observation $i$ and class $k$, and ground truth represented by the binary matrix $y_{ik}$, which is $1$ for the ground truth class $k$ of observation $i$ and $0$ otherwise, the expected cross-entropy loss is given by
\begin{equation}
    \mathrm{XE}=-\frac{1}{N}\sum_{i=1}^N\sum_{k=1}^Ky_{ik}\log(p_{ik})
\end{equation}
and expected Brier score
\begin{equation}
    \mathrm{BS}=\frac{1}{N}\sum_{i=1}^N\frac{1}{K}\sum_{k=1}^K(y_{ik}-p_{ik})^2
\end{equation}
For LC classification, both scores were normalised according to the prior class distribution $q_k$ such that a score of $1$ is no better than a prior class model for all pixels using Figure~\ref{fig:priorPMF}. In this case:
\begin{equation}
\label{eq:XE}
    \mathrm{XE}_\mathrm{norm}=\frac{1}{N}\sum_{i=1}^N\frac{\sum_{k=1}^Ky_{ik}\log(p_{ik})}{\sum_{k=1}^Kq_k\log(q_k)}
\end{equation}
and
\begin{equation}
\label{eq:BS}
    \mathrm{BS}_\mathrm{norm}=\frac{1}{N}\sum_{i=1}^N\frac{\sum_{k=1}^K(y_{ik}-p_{ik})^2}{\sum_{k=1}^Kq_k(1-q_k)}
\end{equation}
\pagebreak
\subsection{Land cover classification}
\label{sec:LCresults}
To fully evaluate the performance of all benchmark on the LC data, we randomly split both the 2020 and 2021 datasets by pixel into a training set and a test set. Due to the large number of pixels on the datasets (189,142 pixels), it was enough\footnote{The change in performance was minimal when taking larger percentages as training data, with only a noticeable increase in computational training time for RFs and NNs.} to evaluate performance by taking small percentages of the data for training the models, namely $0.1\,\%,\,0.5\,\%,\,1\,\%,\,5\,\%$, and $10\,\%$ of pixels. The remaining pixels were held-out for a test dataset, which was fixed for a fair comparison between models. Varying the training set size is important as there are various applications where there is a lack of quality real-world labelled training data, especially when one must also acquire repeated measurement data to capture the input uncertainties. This might be a more realistic scenario if higher quality LC labelled training data was obtained from semi-automatic or manual approaches described in Section~\ref{sec:LCdata}. It also helps evaluate the generalisability of such models to other training datasets of different sizes.

\begin{figure}
    \centering
    \begin{subfigure}{0.4\linewidth}
        \includegraphics[width=\linewidth]{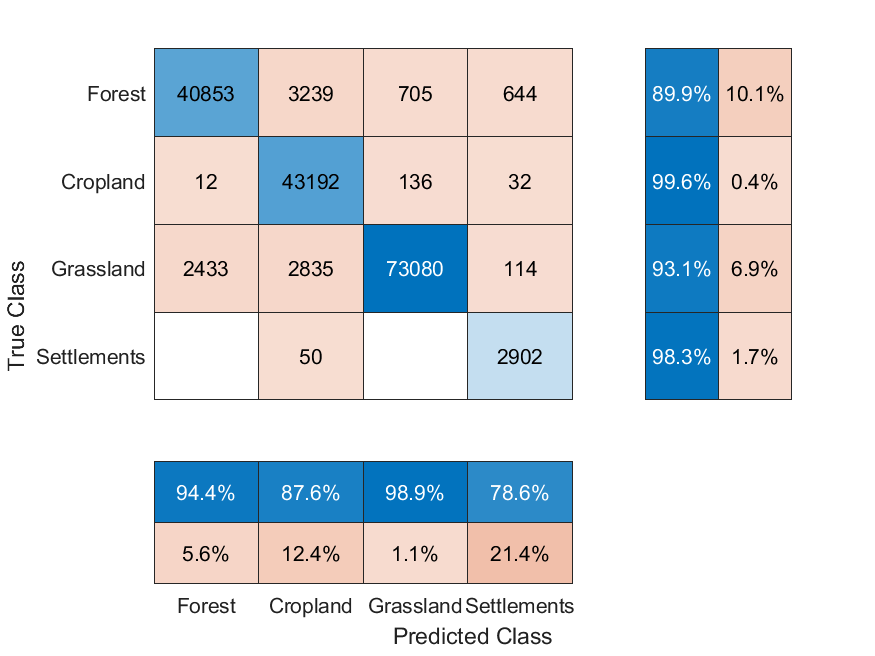}
        \caption{QDA/BQDA}
    \end{subfigure}
    \begin{subfigure}{0.4\linewidth}
        \includegraphics[width=\linewidth]{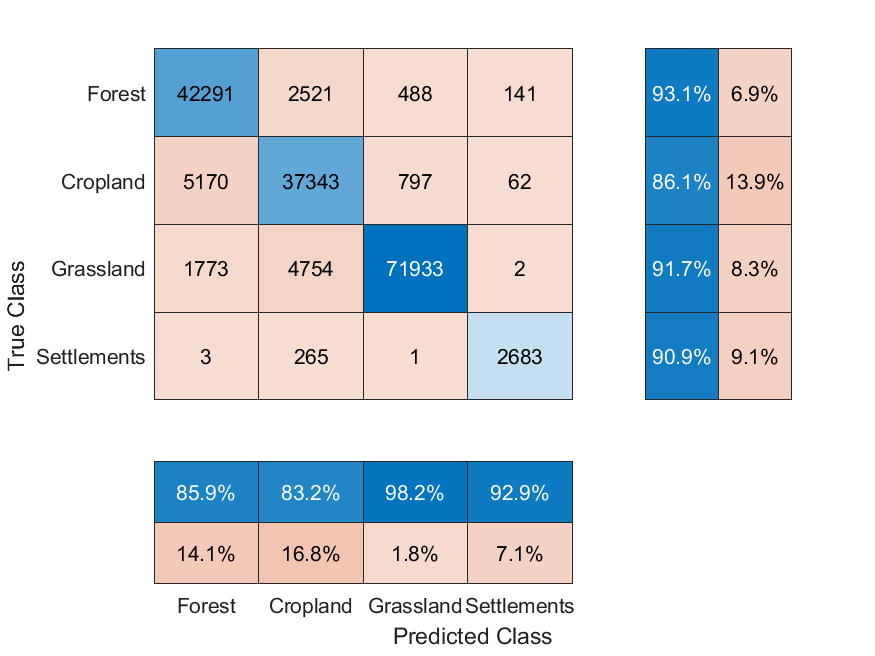}
        \caption{RF}
    \end{subfigure}

    \begin{subfigure}{0.4\linewidth}
        \includegraphics[width=\linewidth]{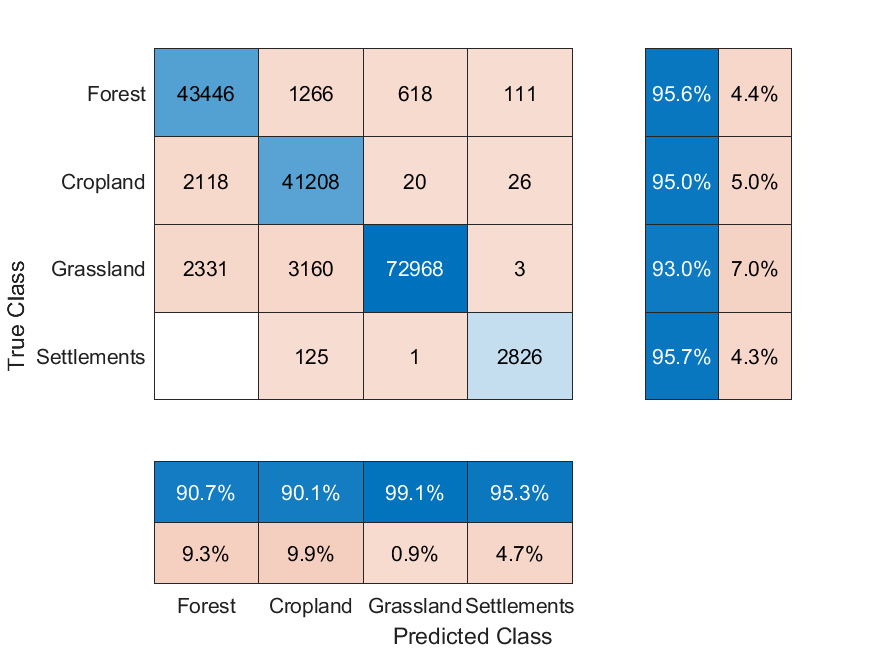}
        \caption{NN}
    \end{subfigure}
    \begin{subfigure}{0.4\linewidth}
        \includegraphics[width=\linewidth]{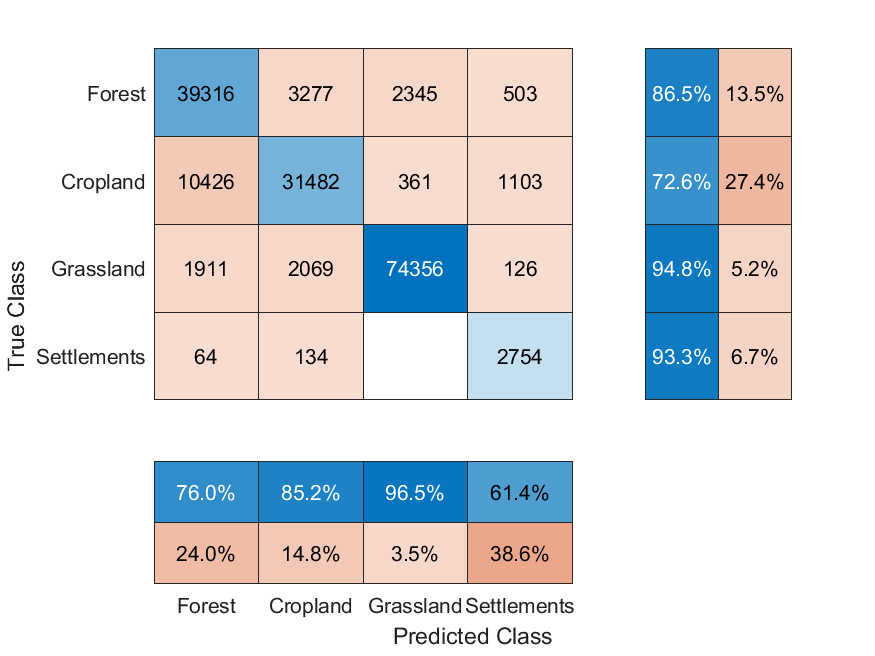}
        \caption{LDA}
    \end{subfigure}   
    \caption{Confusion matrices of model predictions on test data. Models were trained using 10 \% of pixels from year 2020 and tested on the remaining 90 \% of pixels from year 2021. True/false positive rates (bottom rows) and true/false negative rates (right columns) are also included.}
    \label{fig:CM}
\end{figure}

To assess the within-class performance of the classification models, we compute the confusion matrices by taking the maximum predictive probability (mode) as the output class and comparing with the labelled test data classes. Representative results are shown for the case of training on year 2020 and validating on year 2021 in Figure~\ref{fig:CM}. QDA and BQDA have almost identical results for the data sizes chosen, so one confusion matrix is shown for both.

\begin{figure}
    \centering
    \includegraphics[width=0.4\linewidth]{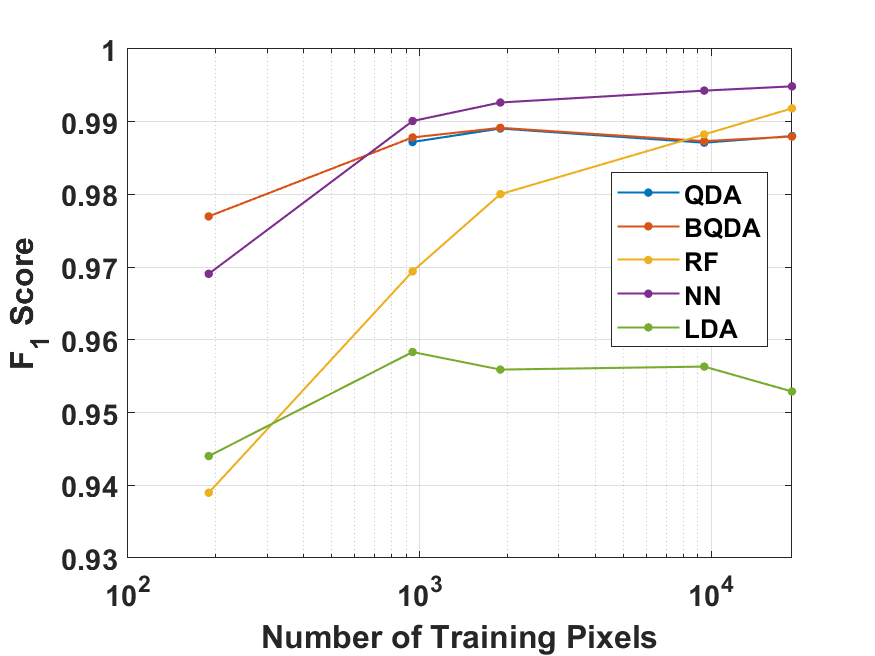}
    \includegraphics[width=0.4\linewidth]{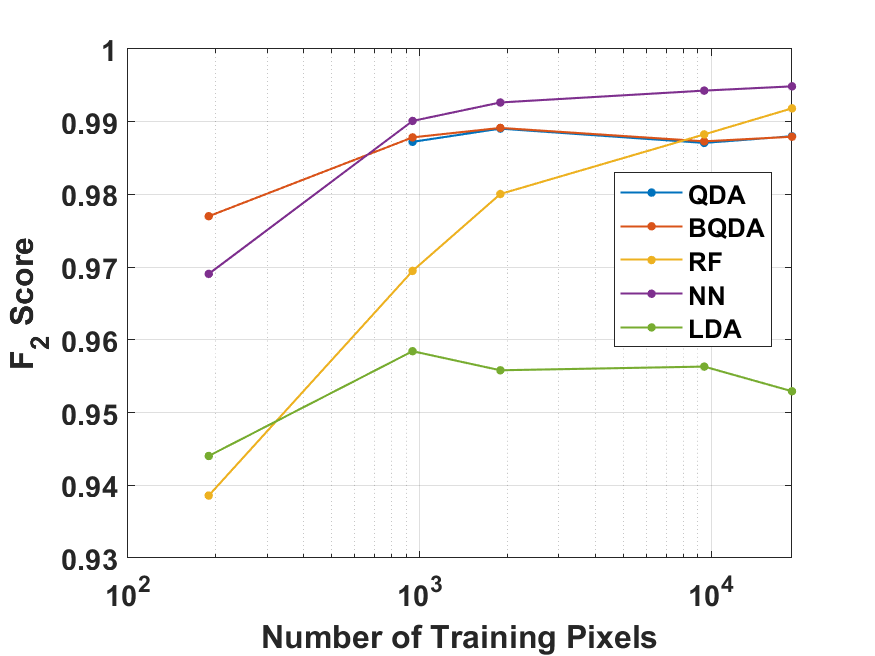}

    \includegraphics[width=0.4\linewidth]{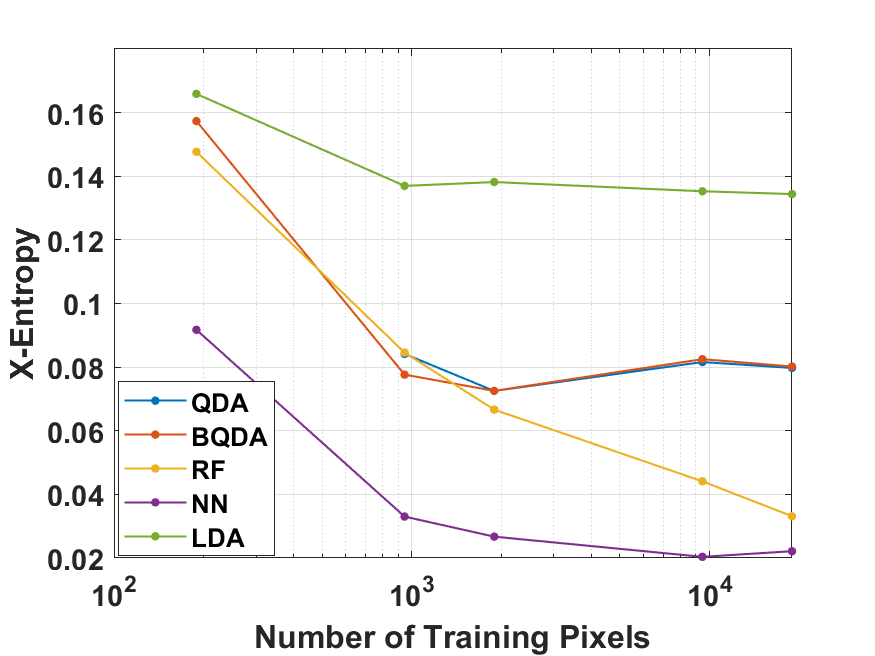}
    \includegraphics[width=0.4\linewidth]{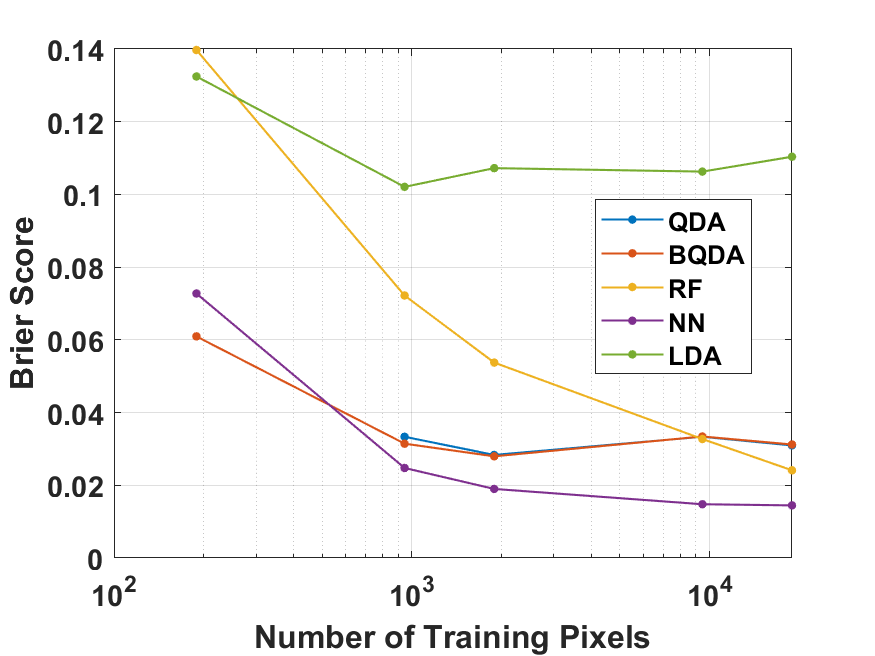}
    \caption{Training year 2020; test year 2020.}
    \label{fig:metrics20_20}
\end{figure}

\begin{figure}
    \centering
    \includegraphics[width=0.4\linewidth]{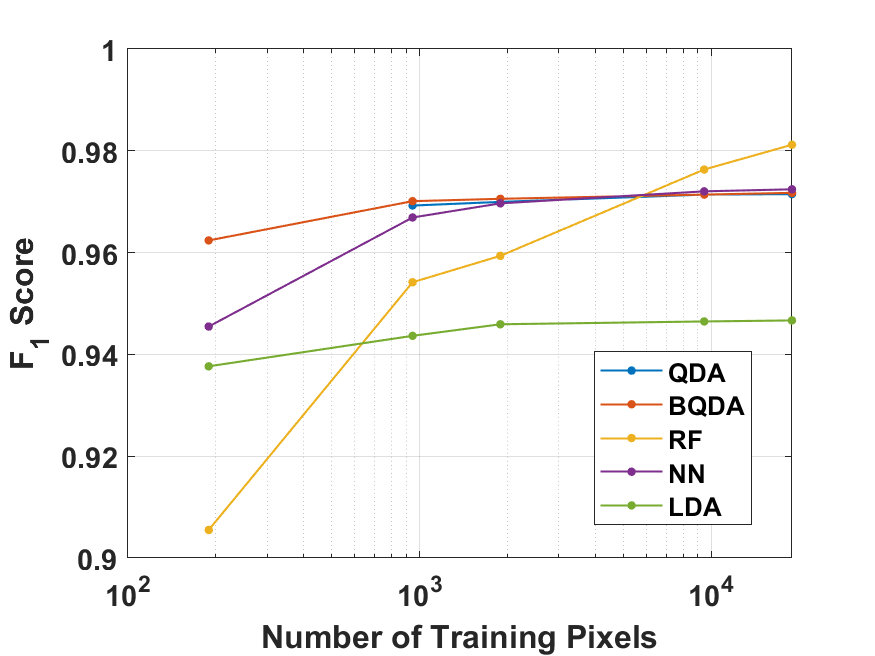}
    \includegraphics[width=0.4\linewidth]{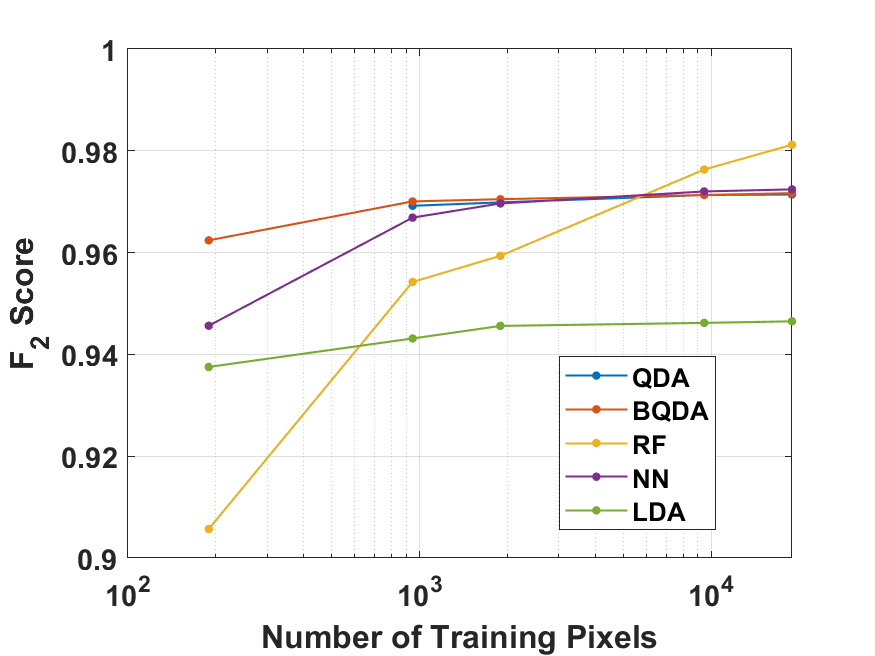}

    \includegraphics[width=0.4\linewidth]{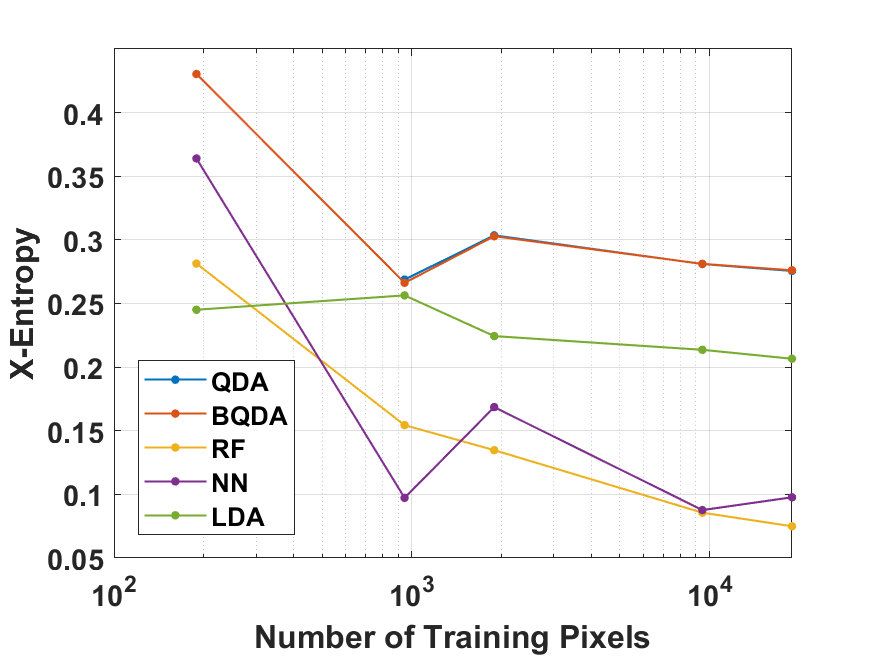}
    \includegraphics[width=0.4\linewidth]{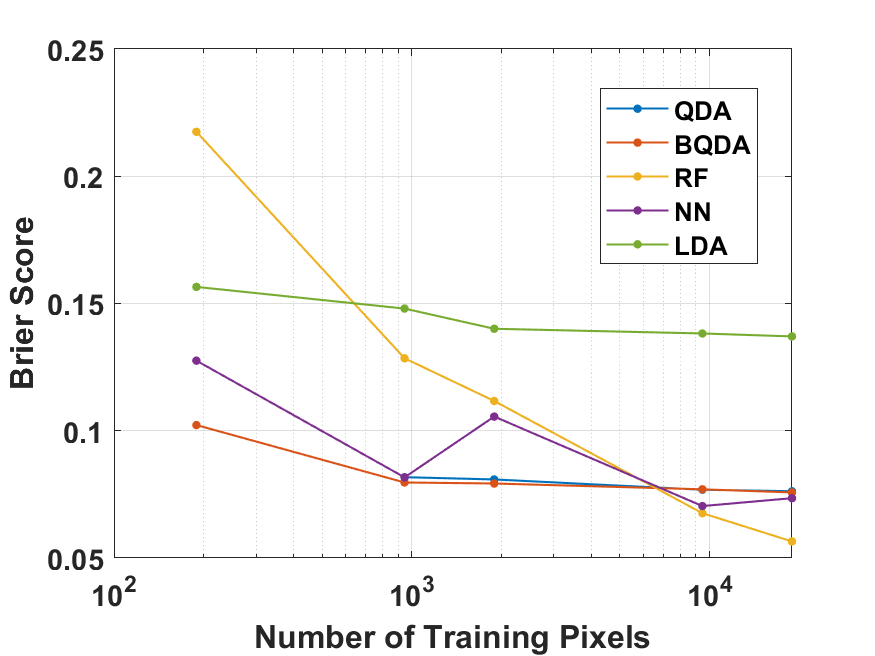}
    \caption{Training year 2021; test year 2021.}
    \label{fig:metrics21_21}
\end{figure}

\begin{figure}
    \centering
    \includegraphics[width=0.4\linewidth]{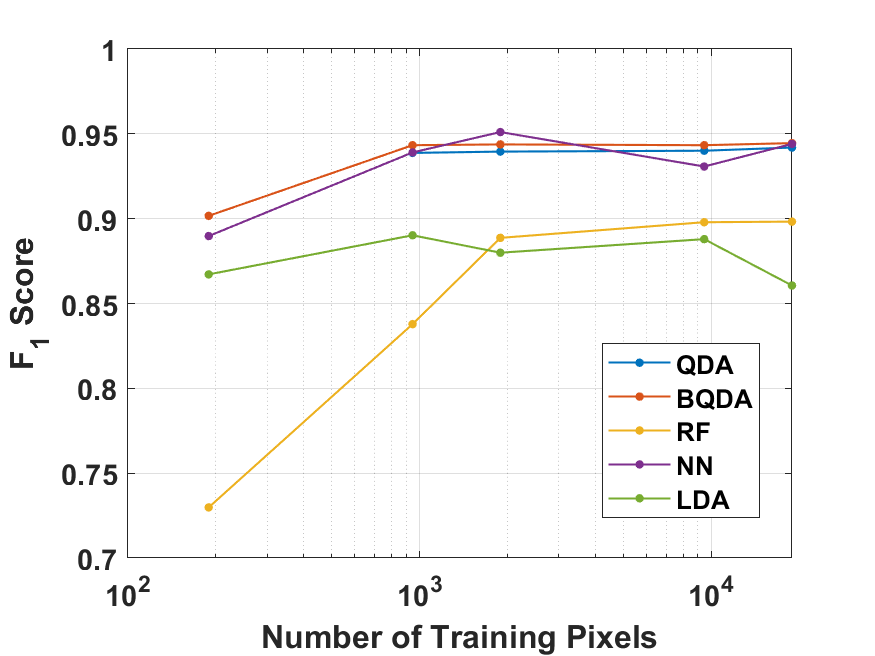}
    \includegraphics[width=0.4\linewidth]{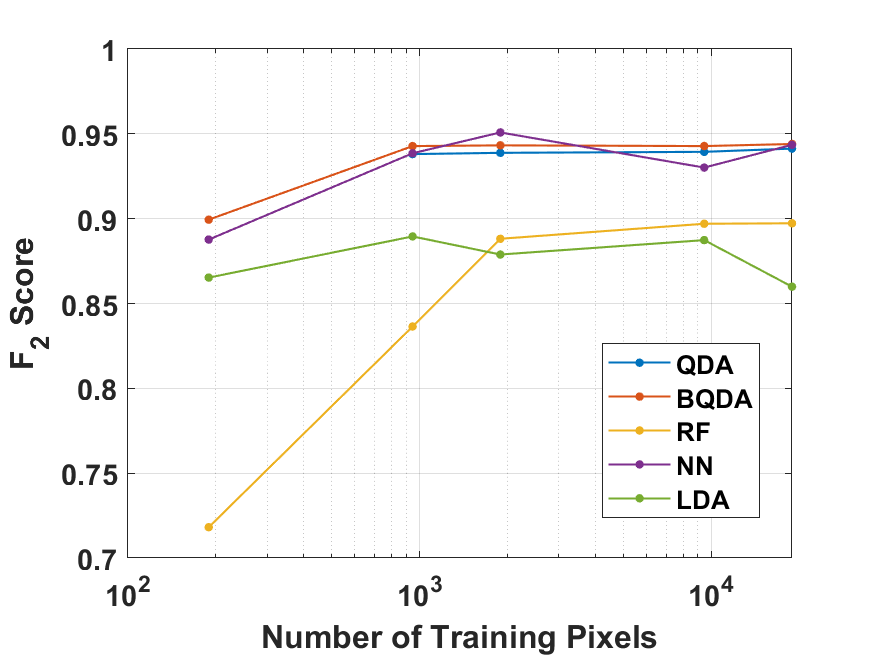}

    \includegraphics[width=0.4\linewidth]{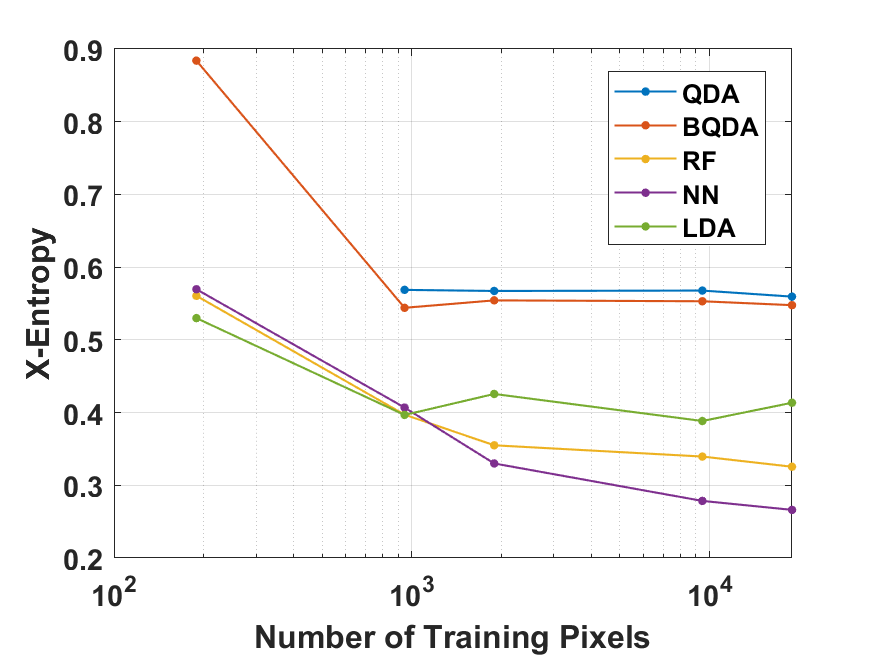}
    \includegraphics[width=0.4\linewidth]{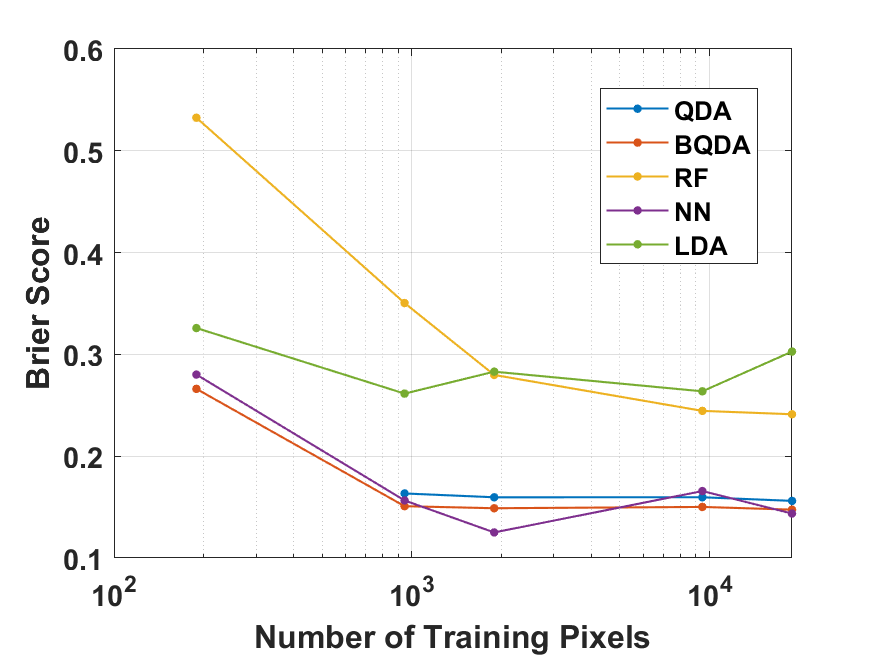}
    \caption{Training year 2020; test year 2021.}
    \label{fig:metrics20_21}
\end{figure}

\begin{figure}
    \centering
    \includegraphics[width=0.4\linewidth]{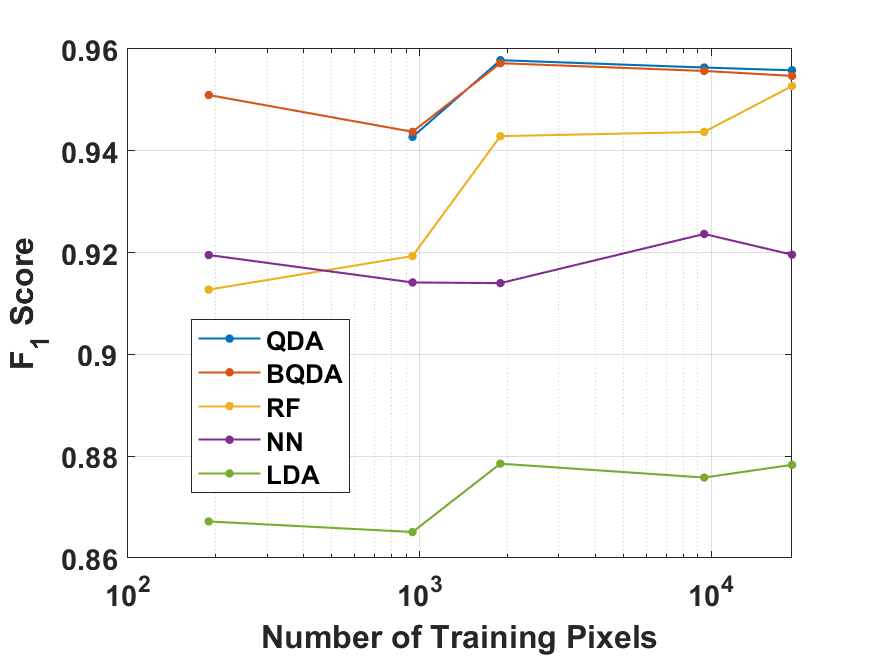}
    \includegraphics[width=0.4\linewidth]{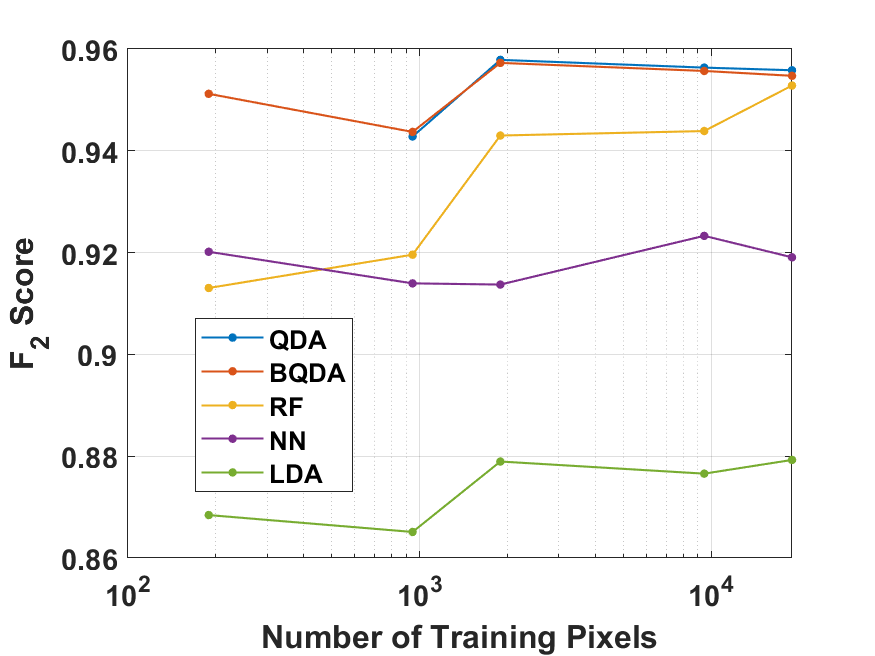}

    \includegraphics[width=0.4\linewidth]{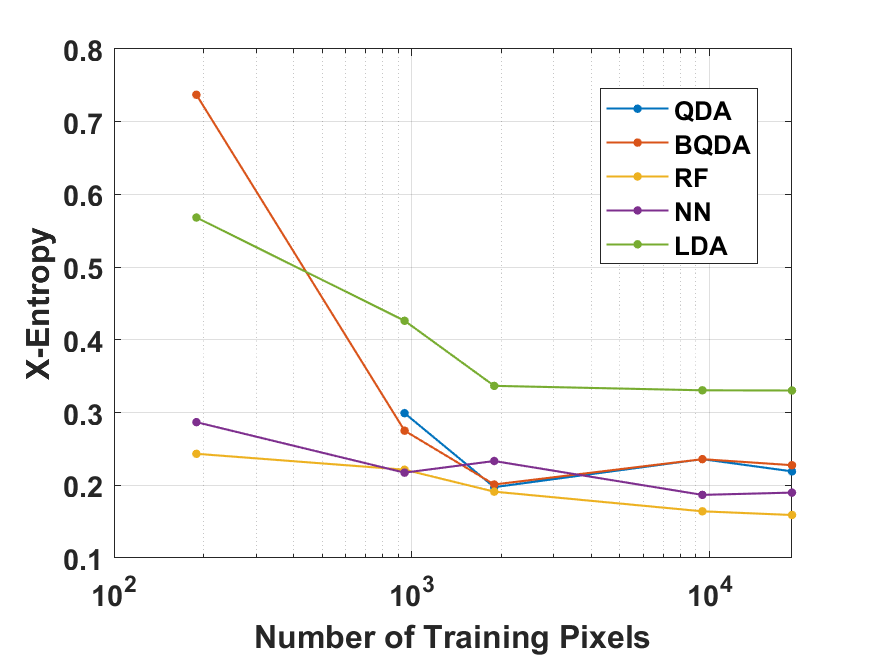}
    \includegraphics[width=0.4\linewidth]{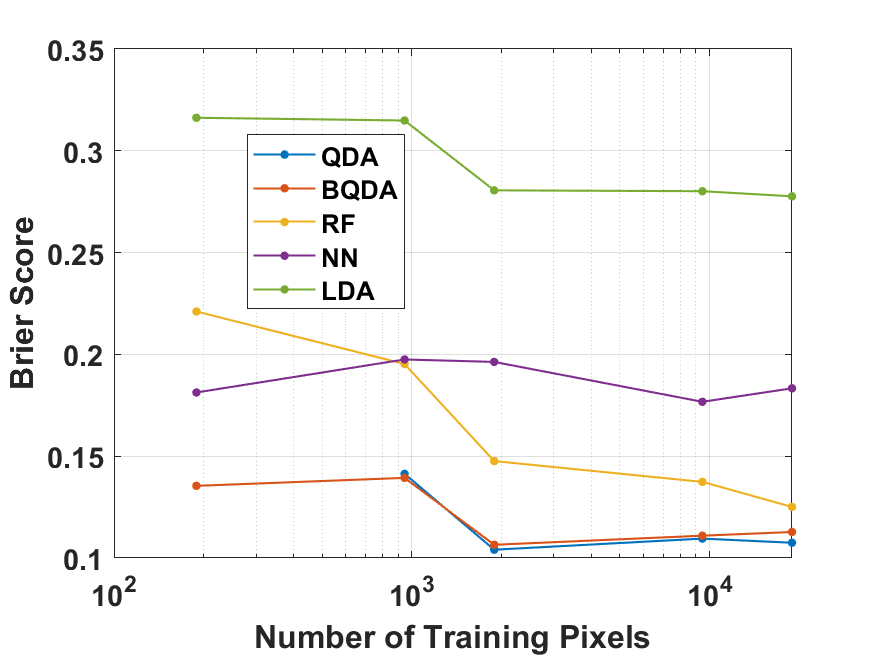}
    \caption{Training year 2021; test year 2020.}
    \label{fig:metrics21_20}
\end{figure}

The validation metrics in expressions~\eqref{eq:XE} and~\eqref{eq:BS} were evaluated on all models trained and tested on the same year (Figures~\ref{fig:metrics20_20} and~\ref{fig:metrics21_21}), as well as models trained and tested on different years (Figures~\ref{fig:metrics20_21} and~\ref{fig:metrics21_20}) to assess model generalisability. 

Addressing within-class performance of all classifiers using Figure~\ref{fig:CM}, we see that all classifiers have a high false positive rate for cropland, which is no surprise considering Figure~\ref{fig:PCA} where cropland is the class with the largest variance in input data. For QDA, forest has a large false negative rate, mostly being predicted as cropland, whereas for NN and RF, cropland has a larger false negate rate and forest has a larger false positive rate than QDA. In other words, NN/RFs tend to over-predict forest and under-predict cropland compared with QDA, but QDA tends to over predict cropland. LDA is the worst performer, as clearly the condition that all classes should have the same covariance biases the model, given the input data distributions for each class in Figure~\ref{fig:PCA}. 

Observing overall performance using Figures~\ref{fig:metrics20_20},\ref{fig:metrics21_21},\ref{fig:metrics20_21},\ref{fig:metrics21_20}, in all cases QDA and BQDA become equivalent in terms of performance metrics as the training data size increases. This is due to the large training data size removing any effect of the prior on the predictive probability outputs (see expression~\eqref{eq:classPostPredQDA}). For smaller training sizes, the QDA performance drops in comparison with BQDA. Indeed for the lowest training size, QDA could not be trained due to the small number of settlement pixels causing the covariance matrix for this class to become singular (non-invertible). The ability to train on small datasets, or datasets with a large class imbalance, is an advantage of the Bayesian approach, which naturally regulates the covariance through the prior in expression~\eqref{eq:priorQDA}, thus maintaining performance at lower numbers of training pixels.

The second point to note is that when the number of training pixels is low ($\leq10^3$), BQDA performs the best in almost all metrics across different years (with the exception of cross-entropy~\eqref{eq:XE}). In the case of cross-entropy loss, the NN was on average the best performer. This is unsurprising since NNs use the cross-entropy as the optimisation function to minimise during training. The poorest performer on average for low numbers of pixels was the random forest, which is the least robust to variations in training data size. This is an important consideration since RFs are widespread in the area of LC classification.

In terms of predictive uncertainty assessment, the only metric that takes account of all class probabilities with an output class predictive distribution is the Brier score~\eqref{eq:BS}. For cases in which training and test years differ (Figures~\ref{fig:metrics20_21} and~\ref{fig:metrics21_20}), BQDA is the lowest and most consistent across training data sizes, making it a good candidate to model the predictive class uncertainties in LC maps with input measurement uncertainties. This is despite that fact that cropland clearly has a non-Gaussian shape, as can be seen in Figure~\ref{fig:PCA}. RFs and NNs perform better when the training and test years are the same.

\begin{figure}
    \centering
    \includegraphics[width=0.6\linewidth]{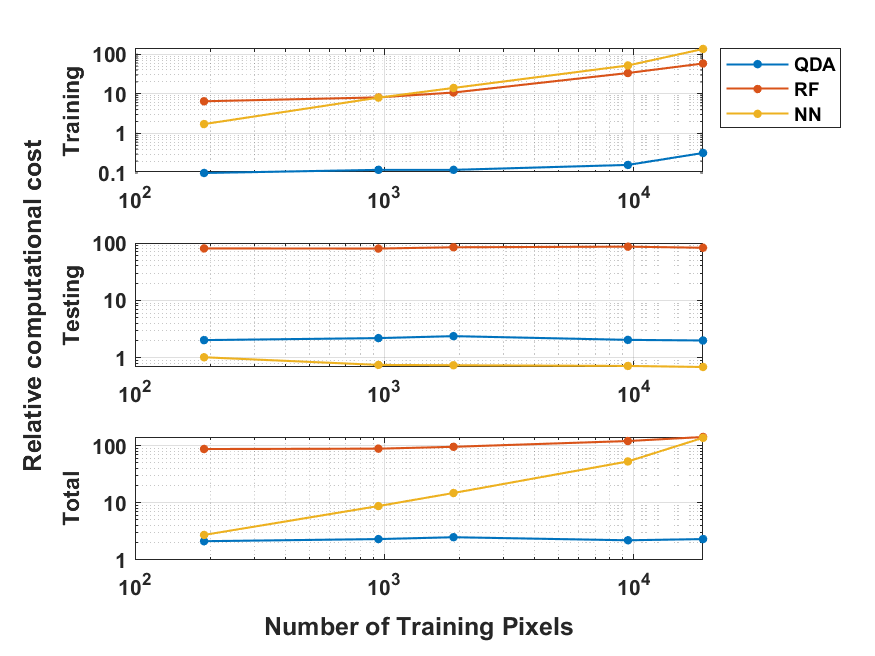}
    \caption{Comparison of computational cost in training and evaluating the selected classification models. BQDA, QDA \& LDA took approximately the same time to implement, so only QDA is shown here. Relative computational cost and number of training pixels are on a log scale.}
    \label{fig:compCost}
\end{figure}

In addition to evaluating model performance on the metrics in Section~\ref{sec:metrics}, we also consider other performance indicators that are important for the use of such models in real-world settings, namely the computational cost. In figure~\ref{fig:compCost}, we plot the relative computational cost of training and evaluating (validating) the models (given the absolute time depends on the particular hardware used to run the code). We see that LDA, QDA, and BQDA are in general more computationally efficient compared with RF and NN. This is no surprise as LDA, QDA, and BQDA have analytic expressions for model training, whereas RFs and NNs do not. This fact also makes the predictive probability outputs of LDA/QDA/BQDA reproducible, given the same training and test data. The equivalent cannot be said for RFs and NNs, as training involves use of random number generation. It is also worth noting that RFs are the most computationally inefficient across different training dataset sizes, which should be taken into account when deploying such models.
\pagebreak
\subsection{Simulation experiments on synthetic data}
\label{sec:sim}
The results described in Section~\ref{sec:LCresults} show good performance of the BQDA model in comparison to benchmark models across a variety of performance metrics on LC classification. However, they are limited by the fixed dataset, and the number of input measurement realisations. To investigate the generalisability of the BQDA model to various sizes and distributions of input noise, we simulate results using synthetic classification data.

We perform experiments upon synthetically-generated classification problems to investigate the impact of input measurement noise on the BQDA model and compare this to the other benchmark models. This input measurement noise could be the result of propagating through a chain of measurement models or through some feature extractor (see Section~\ref{sec:inputUnc}). To generate synthetic data before the addition of measurement noise, multivariate Gaussian distributions are used to define mean and covariance for $p$ input variables and $K$ classes. The $K\times p$ mean value matrix $M$ has i.i.d. entries $M_{ki}\sim\mathcal{N}(0,1)$. The associated $p\times p$ covariance matrix $\Sigma_k$ for each class $k$ is defined using $p\times p$ random matrices $A_k$ with i.i.d. entries,
\begin{equation}
    \Sigma_k=(A_kA_k^\top)/p,\quad\text{where}\quad (A_k)_{ij}\sim\mathcal{N}(0,1),\,i,j=1,\dots p.    
\end{equation}
This ensures the matrices $\Sigma_k$ are symmetric positive semi-definite, while dividing by $p$ ensures the expected values for the covariances $\mathbb{E}[(\Sigma_k)_{ij}]=1$.
By generating class input variables with means and covariances drawn from these distributions we are able produce synthetic classification data with overlapping class distributions, but still separate enough to build models that can achieve a high accuracy before measurement noise is introduced. This allows us to investigate the impact of measurement noise on typical examples faced in the real world. 

For each input variable and for each class, $1,000$ data points are drawn from a multivariate Gaussian distribution with their defined mean and covariance, such that the classes are balanced. We generate datasets with four classes to reflect the LC dataset. Datasets are generated with either $2$, $10$ or $50$ input variables. We conduct experiments to assess the performance of the BQDA model against four benchmark models described in Section~\ref{sec:benchMdl}. Each model is trained and tested on the same 80:20 train/test split of the synthetic data. We compute performance metrics described in Section~\ref{sec:metrics} on the test data for each model.

Datasets are generated with $100$ realisations of measurement noise drawn from either a normal distribution or a uniform distribution. We scale the standard deviation of the measurement noise (known as the``noise factor") from $0.1$ to $1$ with a step size of $0.1$. As the experiments are designed to assess the robustness of the BQDA model to measurement noise, we ensure that the synthetic datasets have class distributions that are separated enough to so that the impact of the measurement noise is observable, but not too well separated that the datasets are unrealistic. This is achieved by generating datasets that produce F$_1$-scores between $0.9$ and $1$ on model predictions, when no measurement noise is added. Thus, minor adjustments are made to the mean matrix $M$ used to generate the synthetic data depending on the number of input variables. For datasets with $2$ input variables, the values drawn from a Gaussian distribution used to define the mean of the input variables are multiplied by the number of input variables squared. For the datasets with $50$ input variables, the values used to define the mean values was divided by $2$. For datasets defined by $10$ input variables, the mean matrix is not modified.

\begin{figure}
    \centering
    \includegraphics[scale=0.5]{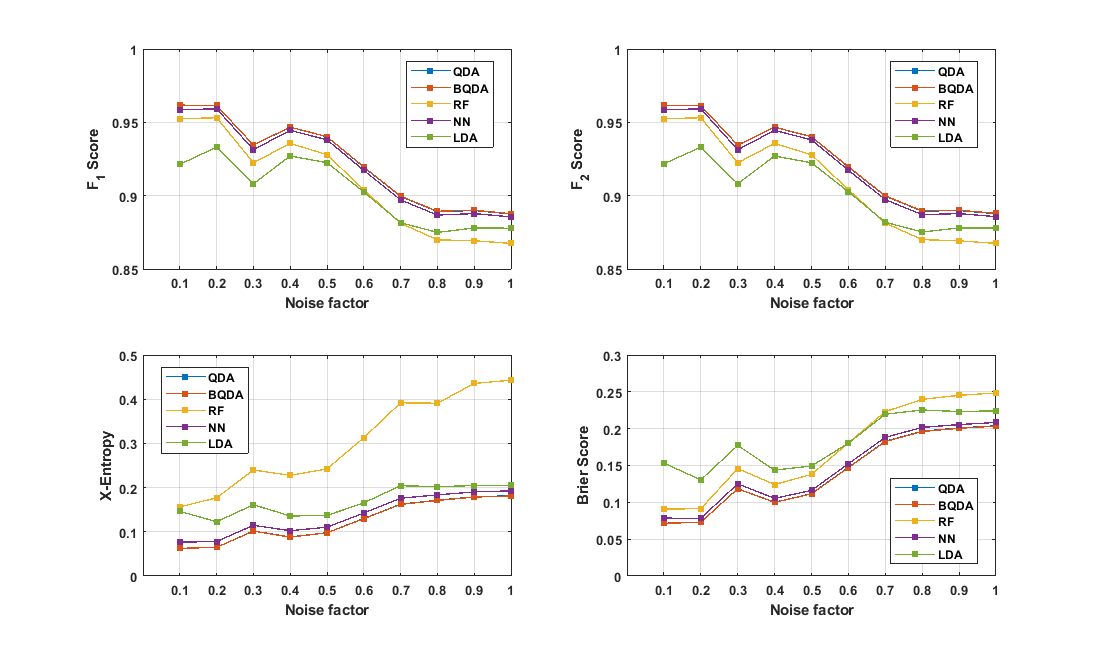}
    \caption{Metric scores for five models trained and tested on synthetic datasets with two variables and varying strength of Gaussian distributed measurement noise.}
    \label{fig:SimG2F}
\end{figure}

\begin{figure}
    \centering
    \includegraphics[scale=0.5]{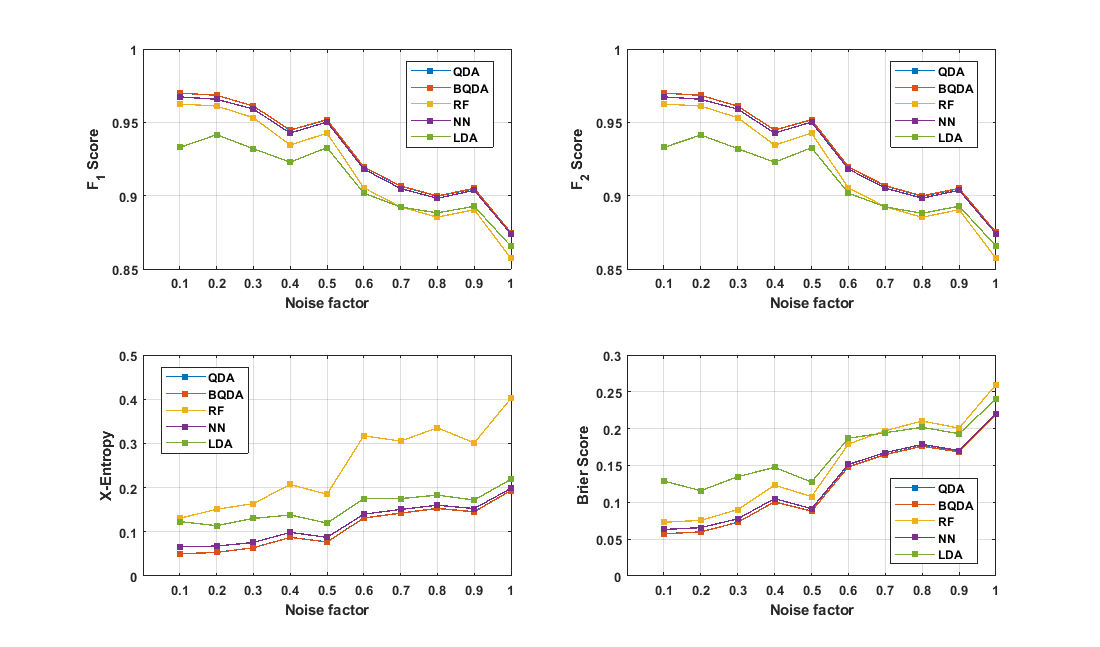}
    \caption{Metric scores for five models trained and tested on synthetic datasets with two variables and varying strength of uniform distributed measurement noise.}
    \label{fig:SimU2F}
\end{figure}

\begin{figure}
    \centering
    \includegraphics[scale=0.5]{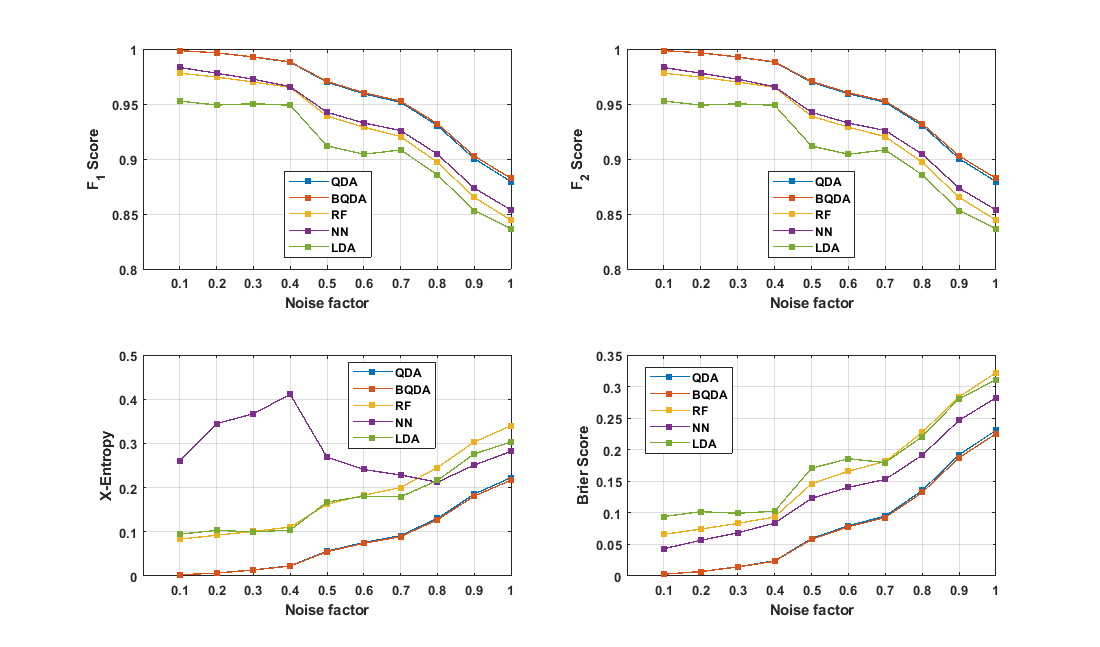}
    \caption{Metric scores for five models trained and tested on synthetic datasets with ten variables and varying strength of Gaussian distributed measurement noise.}
    \label{fig:SimG10F}
\end{figure}

\begin{figure}
    \centering
    \includegraphics[scale=0.5]{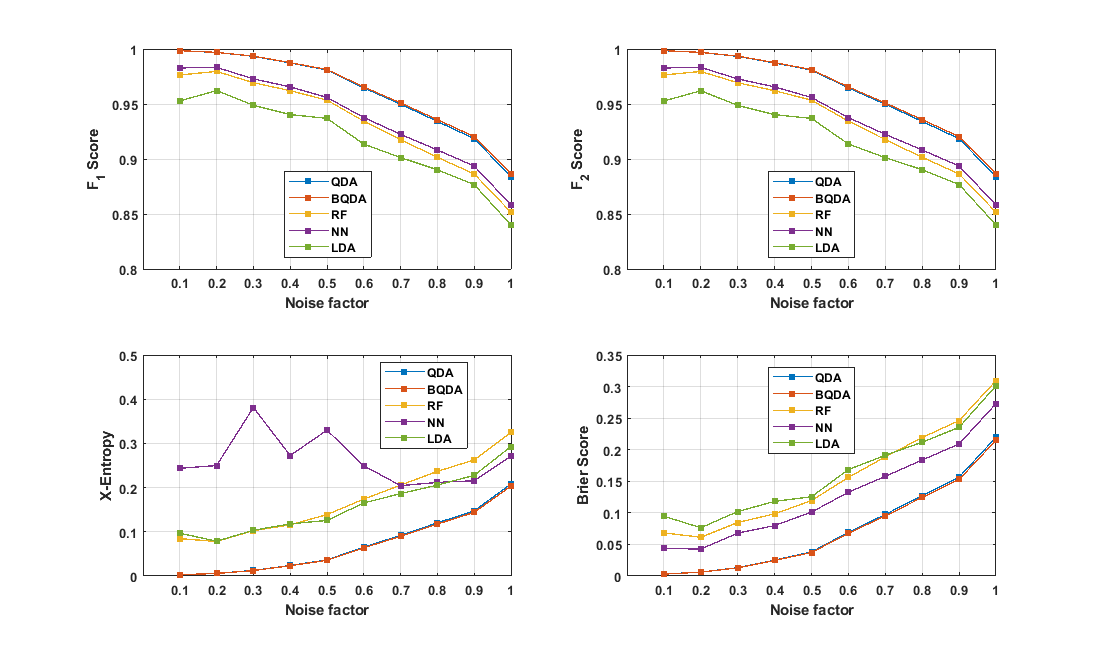}
    \caption{Metric scores for five models trained and tested on synthetic datasets with ten variables and varying strength of uniform distributed measurement noise.}
    \label{fig:SimU10F}
\end{figure}

\begin{figure}
    \centering
    \includegraphics[scale=0.5]{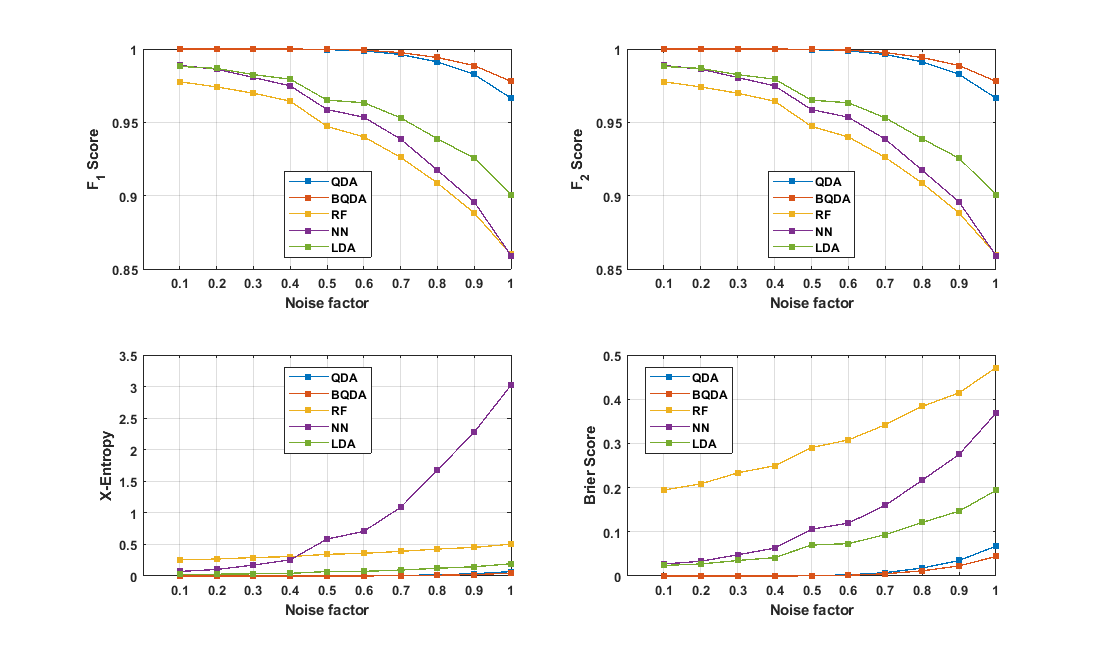}
    \caption{Metric scores for five models trained and tested on synthetic datasets with fifty variables and varying strength of Gaussian distributed measurement noise.}
    \label{fig:SimG50F}
\end{figure}

\begin{figure}
    \centering
    \includegraphics[scale=0.5]{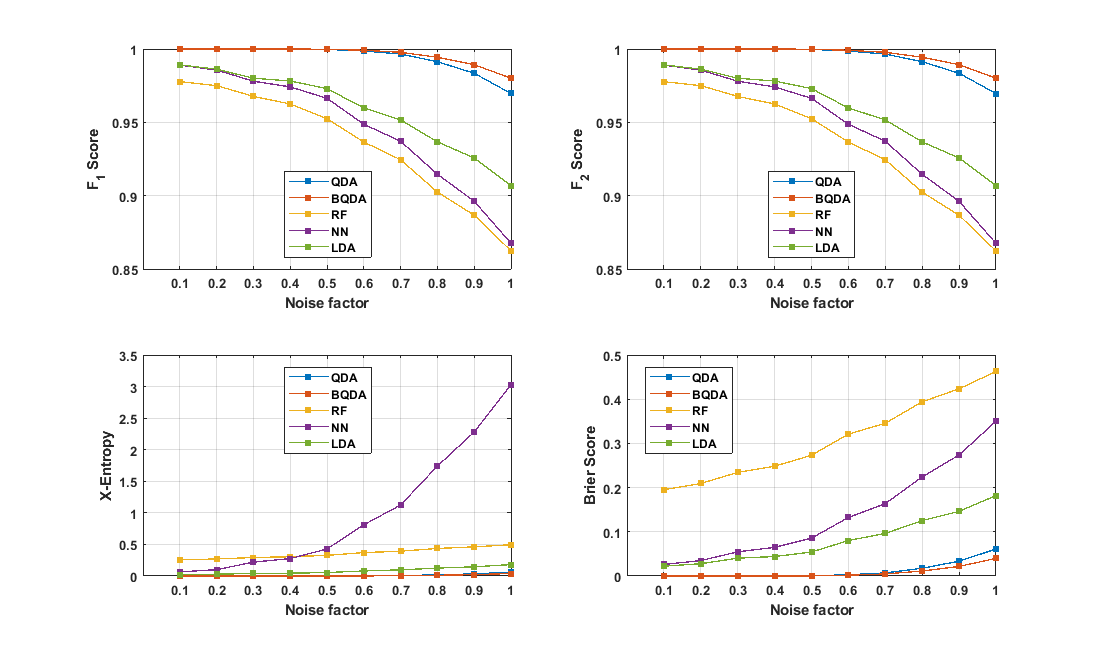}
    \caption{Metric scores for five models trained and tested on synthetic datasets with fifty variables and varying strength of uniform distributed measurement noise.}
    \label{fig:SimU50F}
\end{figure}

The metric scores presented in Figures \ref{fig:SimG2F} and \ref{fig:SimU2F} are the average metric score over $60$ different datasets with $100$ repeated measures. The metric scores presented in Figures \ref{fig:SimU10F} to \ref{fig:SimU50F} are the average metric score over $30$ different datasets with $100$ repeated measurements. We observe for synthetic datasets consisting of two variables shown in Figure \ref{fig:SimG2F} and Figure \ref{fig:SimU2F}, all models have worse metrics as the noise factor is increased, though significant fluctuations are observed most likely due to the small number of features. All models generally follow the same trend and there is no significant changes in the models performance as the noise factor increases. Interestingly, the neural network model performs just as well as the BQDA model. 

For synthetic datasets consisting of ten variables shown in Figure \ref{fig:SimG10F} and Figure \ref{fig:SimU10F}, we observe all five models showing a worsening of metric scores as the noise factor is increased. Once again, we observe all models to follow the same trend as the noise factor increases. However, we observe for ten variables, the BQDA model always achieves a better metric score. 

For synthetic datasets consisting of fifty variables shown in Figure \ref{fig:SimG50F} and Figure \ref{fig:SimU50F} we observe interesting changes in behaviour between the models. For each model, we observe worsening in metric scores as the noise factor is increased. However, the BQDA model not only achieves a better metric score than the other models but it also decays at a slower rate compared to the other models. Here we see the advantage of a Bayesian approach, as the prior (see expression~\eqref{eq:priorQDA}) acts as a regulariser to avoid the over-fitting seen in the frequentist QDA benchmark model. 

For all numbers of features, we do not observe significant changes in behaviour for synthetic datasets with Gaussian measurement noise compared to uniform measurement noise. We have also conducted similar experiments with no covariance between variables and found this to return very similar results to those presented here.

\pagebreak
\section{Conclusion}
\label{sec:con}
We present an uncertainty-aware Bayesian ML classification framework for handling input measurement uncertainty. We focus on parametric generative ML classifiers, and specifically Bayesian quadratic discriminant analysis. We take as our case study the production of uncertainty-aware land cover maps, for which input measurement uncertainty plays an important role. We compare the performance of Bayesian QDA against other well known classification models used in LC classification, that do not explicitly take account of such input uncertainty. We address performance on LC maps and synthetic datasets in terms of classification outputs, class probabilities, and computational cost. We find that such Bayesian models are more interpretable and computationally efficient, whilst maintaining predictive performance of class probability outputs across test datasets of different years and sizes, as well as for Gaussian and uniform measurement noise with varying strengths. This is despite the fact that Bayesian QDA is highly constrained by the choice of input measurement model and class conditional distributions (both multivariate Gaussian), compared with more complex models such as RFs and NNs. The results show the utility of a Bayesian approach for incorporating measurement input uncertainty in machine learning classification models. 

One limitation of this work is that class labelling uncertainty was not considered, but is an important consideration in many applications, such as might occur in LC mapping when a pixel contains multiple classes. Having an explicit Bayesian framework allows the extension of the BQDA model to include such uncertainties, which we leave for future work. We have also restricted our attention to measurement noise that is the result of propagating through a chain of measurement models, or feature extractor, where the input measurement uncertainty distribution is unknown, and needs to be learned. We leave the case of known input measurement distributions for future work.

\pagebreak
\section*{Acknowledgements}
The authors would like to thank Peter Harris and Andrew Thompson (NPL) for very useful comments and review of this article. We would also like to thank Paul Green, and Spencer Thomas (NPL) for their feedback and advice.

\section*{Funding}
This work was funded by the UK Government's Department for Science, Innovation and Technology (DSIT) through the UK's National Measurement System (NMS) programmes. Anna Pustogvar’s work was also supported by a College of Science and Engineering (CSE) Scholarship from the University of Leicester.

\section*{Data}
The original data used for LC classification can be found at~\cite{Morton2020a,Morton2020b,Morton2020c,Morton2020d}.
\pagebreak
\printbibliography
\pagebreak
\appendix
\section{Bayesian generative classification models}
\label{ap:bayes}
\subsection{Likelihood}
\label{sec:like}
As with most probabilistic models, we begin with a likelihood of the training data given model parameters. We take the usual assumption in ML classification models that observations are independent, even though this may not be true in reality, then the likelihood of generating data $\mathcal{D}=\{\mathbf{x}_i,y_i\}_{i=1}^N$ from a classification model with parameters $\pmb{\theta}$ and $\pmb{\phi}$ is given by
\begin{equation}
    p(\mathcal{D}|\pmb{\theta},\pmb{\phi})=\prod_{i=1}^Np(\mathbf{x}_i,y_i|\pmb{\theta},\pmb{\phi}),
\end{equation}
where we assume independence of parameters $\pmb{\theta}$ and $\pmb{\phi}$ which are model parameters for the input and output data distributions respectively. 

For a generative classifier, the input data distribution $p(\mathbf{x}_i|y_i,\pmb{\theta})$ is explicitly modelled for a given class $c_k$, thus we can rewrite the likelihood as
\begin{equation}
    p(\mathcal{D}|\pmb{\theta},\pmb{\phi})=\prod_{i=1}^Np(\mathbf{x}_i|y_i,\pmb{\theta})\times\prod_{i=1}^Np(y_i|\pmb{\phi}).
\end{equation}
The variable $y_i$ is categorical, and can be modelled by a categorical distribution without loss of generality
\begin{equation}
    p(y_i|\pmb{\phi})=\prod_{k=1}^K\phi_k^{\mathbbm{1}(y_i=c_k)},
\end{equation}
where $\phi_k=P(y=c_k|\pmb{\phi})$ is the probability of an observation $y$ belonging to class $c_k$, and $\mathbbm{1}(y_i=c_k)$ is the indicator function that is $1$ when $y_i=c_k$, and $0$ otherwise. 

If we assume specific model parameters $\pmb{\theta}_k$ for each independent class $c_k$, then the likelihood can be rewritten as
\begin{align}
    \label{eq:like1}
    p(\mathcal{D}|\pmb{\theta},\pmb{\phi})&=\prod_{i=1}^N\prod_{k=1}^Kp(\mathbf{x}_i|\pmb{\theta}_k)^{\mathbbm{1}(y_i=c_k)}\times\prod_{i=1}^N\prod_{k=1}^K\phi_k^{\mathbbm{1}(y_i=c_k)}\nonumber\\
    &=\prod_{k=1}^K\phi_k^{N_k}\prod_{i|y_i=c_k}^{N_k}p(\mathbf{x}_i|\pmb{\theta}_k)
\end{align}
where $N_k$ is the number of training data points in class $c_k$, and so $\sum_{k=1}^KN_k=N$.

\subsection{Priors on model parameters}
In a Bayesian approach, model parameters come with prior distributions, which express prior knowledge (or lack of) about the values of the model parameters. Given the independence assumption, we can express the prior probability on the model parameters as
\begin{equation}
    p(\pmb{\theta},\pmb{\phi})=p(\pmb{\theta})p(\pmb{\phi}).
\end{equation}
For the model class distribution $p(y_i|\pmb{\phi}$), we can use the conjugate prior for the categorical distribution, which is the Dirichlet distribution $\pmb{\phi}\sim\mathrm{Dir}(\alpha_1,\dots,\alpha_K)$, written as
\begin{equation}\label{eq:dirichlet}
	p(\pmb{\phi}|\pmb{\alpha})=\frac{1}{B(\pmb{\alpha})}\prod_{k=1}^{K}\phi_k^{\alpha_k-1},\quad\pmb{\phi}\geq\mathbf{0},\,\displaystyle\sum_{k=1}^{K}\phi_k=1,
\end{equation}
where $\pmb{\alpha}=[\alpha_1,\dots,\alpha_K]^\top\geq\mathbf{0}$ are the prior hyperparameters and $B(\pmb{\alpha})$ is the multivariate Beta function. 

The expected value of this distribution is
\begin{equation}
	\mathbb{E}[\pmb{\phi}]=\frac{\pmb{\alpha}}{\alpha_0},\;\;\;\;\mathrm{where}\;\alpha_0=\sum_{k=1}^{K}\alpha_k.
	\label{eq:DirProp}
\end{equation}
So the hyperparameters $\alpha_k$ represent the prior weights of each class. Thus we can write the prior distribution as
\begin{equation}
\label{eq:prior}
    p(\pmb{\theta},\pmb{\phi}|\pmb{\alpha},\pmb{\beta})=\prod_{k=1}^K\phi^{\alpha_k-1}p(\pmb{\theta}_k|\pmb{\beta}_k),
\end{equation}
where $\pmb{\theta}_k$ are the input data model parameters for each class $c_k$, and $\pmb{\beta}_k$ are the hyperparameters of the prior distribution.

The form of the priors $p(\pmb{\theta}_k|\pmb{\beta}_k)$ depend on the choice of model. The conjugate prior for Bayesian QDA is given in expression~\eqref{eq:priorQDA}.
\subsection{Posterior distribution}
The posterior distribution of the classification model parameters is given by
\begin{align}
\label{eq:post1}
    p(\pmb{\theta},\pmb{\phi}|\mathcal{D},\pmb{\alpha},\pmb{\beta})&=\frac{p(\mathcal{D}|\pmb{\theta},\pmb{\phi})p(\pmb{\theta},\pmb{\phi}|\pmb{\alpha},\pmb{\beta})}{p(\mathcal{D}|\pmb{\alpha},\pmb{\beta})}\nonumber\\
    &=p(\pmb{\phi}|\mathcal{D},\pmb{\alpha})p(\pmb{\theta}|\mathcal{D},\pmb{\beta})\nonumber\\
    &=\prod_{k=1}^Kp(\phi_k|\mathcal{D}_k,\alpha_k)p(\pmb{\theta}_k|\mathcal{D}_k,\pmb{\beta}_k)
\end{align}
where we assumed independence of $\pmb{\theta}$ and $\pmb{\phi}$, and $\mathcal{D}_k=\{(\mathbf{x}_i,y_i)|y_i=c_k\}$ is the training data for class $c_k$. 

The marginal posteriors for each class $c_k$ are given by
\begin{equation}
    \label{eq:margPostPhi}
    p(\phi_k|\mathcal{D}_k,\alpha_k)\propto\phi_k^{\alpha_k+N_k-1}
\end{equation}
and
\begin{align}
\label{eq:margPostTheta}
p(\pmb{\theta}_k|\mathcal{D}_k,\pmb{\beta}_k)&=\frac{p(\mathcal{D}_k|\pmb{\theta}_k)p(\pmb{\theta}_k|\pmb{\beta}_k)}{p(\mathcal{D}_k|\pmb{\beta}_k)}\nonumber\\
&=\frac{\prod_{i|k}p(\mathbf{x}_i|\pmb{\theta}_k)p(\pmb{\theta}_k|\pmb{\beta}_k)}{\prod_{i|k}\int d\pmb{\theta}_kp(\mathbf{x}_i|\pmb{\theta}_k)p(\pmb{\theta}_k|\pmb{\beta}_k)},
\end{align}
where the condition $i|k$ is compact notation for all input training measurements $\mathbf{x}_i$ with labelled class $k$, i.e. $y_i=c_k$.
\subsection{Posterior predictive distributions}
Using the posterior~\eqref{eq:margPostPhi}, we have
\begin{align}
\label{eq:postPredY}
    P(\Tilde{y}=c_k|\mathcal{D},\pmb{\alpha})&=\int d\pmb{\phi}\,p(\Tilde{y}|\phi_k)p(\pmb{\phi}|\mathcal{D},\pmb{\alpha})\nonumber\\
    &=\int d\phi_k\phi_kp(\phi_k|\mathcal{D}_k,\alpha_k)\prod_{m\neq k}\int d\phi_m p(\phi_m|\mathcal{D}_m,\alpha_m)\nonumber\\
    &=\mathbb{E}[\phi_k|\mathcal{D}_k,\alpha_k]\propto N_k+\alpha_k,
\end{align}
where for the final line we have used the expectation value of the Dirichlet distribution given in expression~\eqref{eq:DirProp}. Similarly, using expression~\eqref{eq:margPostTheta} we have
\begin{align}
    \label{eq:predPostX1}
    p(\Tilde{\mathbf{x}}|\Tilde{y}=c_k,\mathcal{D},\pmb{\beta})&=\int d\pmb{\theta}\,p(\Tilde{\mathbf{x}}|\pmb{\theta}_k)p(\pmb{\theta}|\mathcal{D},\pmb{\beta})\nonumber\\
    &=\int d\pmb{\theta}_kp(\Tilde{\mathbf{x}}|\pmb{\theta}_k)p(\pmb{\theta}_k|\mathcal{D}_k,\pmb{\beta}_k)\prod_{m\neq k}\int d\pmb{\theta}_mp(\pmb{\theta}_m|\mathcal{D}_m,\pmb{\beta}_m)\nonumber\\
    &=\int d\pmb{\theta}_kp(\Tilde{\mathbf{x}}|\pmb{\theta}_k)p(\pmb{\theta}_k|\mathcal{D}_k,\pmb{\beta}_k)\nonumber\\
    &=\int d\pmb{\theta}_kp(\Tilde{\mathbf{x}}|\pmb{\theta}_k)\prod_{i|k}\frac{p(\mathbf{x}_i|\pmb{\theta}_k)p(\pmb{\theta}_k|\pmb{\beta}_k)}{p(\mathbf{x}_i|\pmb{\beta}_k)}.
\end{align}
\pagebreak
\section{Bayesian inference for multivariate Gaussian distributions}
\label{ap:MVG}
The posterior predictive distribution of a multivariate Gaussian distribution $\mathcal{N}(\mathbf{x}|\pmb{\mu},\Sigma)$, where both the mean $\pmb{\mu}$ and covariance $\Sigma$ are unknown, is given by
\begin{equation}
    \label{eq:postPredQDA}
    p(\Tilde{\mathbf{x}}|\mathcal{D},\pmb{\beta})=\int d\pmb{\mu} d\Sigma\,\mathcal{N}(\Tilde{\mathbf{x}}|\pmb{\mu},\Sigma)p(\pmb{\mu},\Sigma|\mathcal{D},\pmb{\beta}),
\end{equation}
where
\begin{equation}
    p(\pmb{\mu},\Sigma|\mathcal{D},\pmb{\beta})=\prod_i\frac{\mathcal{N}(\mathbf{x}_i|\pmb{\mu},\Sigma)p(\pmb{\mu},\Sigma|\pmb{\beta})}{p(\mathbf{x}_i|\pmb{\beta})},
\end{equation}
is the posterior for the multivariate Gaussian distribution using expression~\eqref{eq:predPostX1}, $p(\pmb{\mu},\Sigma|\pmb{\beta})$ is the prior distribution, $\pmb{\beta}$ are the prior hyperparameters, and $\mathcal{D}=\{\mathbf{x}_i\}_{i=1}^N$ is the input training data. 
\subsection{Conjugate prior}
Expression~\eqref{eq:postPredQDA} has an analytic result if we choose $p(\pmb{\mu},\Sigma|\pmb{\beta})$ as the conjugate prior of the multivariate Gaussian distribution. This is known as the Normal-Inverse-Wishart (NIW) distribution
\begin{equation}
    p(\pmb{\mu},\Sigma|\pmb{\beta})=\mathrm{NIW}(\pmb{\mu}_0,\lambda,\pmb{\Psi},\nu),
\end{equation}
where
\begin{align}
\label{eq:NIW}
    \mathrm{NIW}&(\pmb{\mu},\Sigma|\pmb{\mu}_0,\lambda,\pmb{\Psi},\nu)\nonumber\\
    &=\mathcal{N}\left(\pmb{\mu}\Big|\pmb{\mu}_0,\frac{\Sigma}{\lambda}\right)\mathcal{W}^{-1}(\Sigma|\pmb{\Psi},\nu)\nonumber\\
    &=\frac{\lambda^{p/2}|\pmb{\Psi}|^{\nu /
    2}|\Sigma|^{-\frac{\nu + p + 2}{2}}}{(2\pi)^{p/2}2^{\frac{\nu p}{2}}\Gamma_p(\frac{\nu}{2})}\mathrm{exp}\left\{-\frac{1}{2}\mathrm{Tr}(\Psi\Sigma^{-1})-\frac{\lambda}{2}(\pmb{\mu}-\pmb{\mu}_0)^\top\Sigma^{-1}(\pmb{\mu} - \pmb{\mu}_0) \right\},
\end{align}
with expected values $\mathbb{E}_{\mathrm{NIW}}[\pmb{\mu}]=\pmb{\mu}_0$,  $\mathbb{E}_{\mathrm{NIW}}[\Sigma]=\frac{\pmb{\Psi}}{\nu-p-1}$, and $\mathcal{W}^{-1}$ is the inverse Wishart distribution.
\subsection{Posterior distribution}
The posterior distribution for $\mathcal{N}(\mathbf{x}_i|\pmb{\mu},\Sigma)$ will then also be NIW:
\begin{equation}
    \label{eq:postQDA}
    p(\pmb{\mu},\Sigma|\mathcal{D},\pmb{\beta})=\mathrm{NIW}(\pmb{\mu}_N,\lambda_N,\pmb{\Psi}_N,\nu_N),
\end{equation}
where $N$ is the number of training data observations, and
\begin{align*}
    \pmb{\mu}_N&=\frac{\lambda\pmb{\mu}_0+N\Bar{\mathbf{x}}}{\lambda_N}\\
    \lambda_N&=\lambda+N\\
    \nu_N&=\nu+N\\
    \pmb{\Psi}_N&=\pmb{\Psi}+\mathbf{S}+\frac{\lambda N}{\lambda_N}(\Bar{\mathbf{x}}-\pmb{\mu}_0)(\Bar{\mathbf{x}}-\pmb{\mu}_0)^\top\\
    \mathbf{S}&=\sum_{i=1}^N(\mathbf{x}_i-\Bar{\mathbf{x}})(\mathbf{x}_i-\Bar{\mathbf{x}})^\top.
\end{align*}
\subsection{Posterior predictive distribution}
Using the expression for the posterior~\eqref{eq:postQDA}, the posterior predictive distribution~\eqref{eq:postPredQDA} is a multivariate Student-$t$ distribution~\cite[Chapter~4]{murphy_machine_2012}
\begin{equation}
    \label{eq:postPredTheta}
    p(\Tilde{\mathbf{x}}|\mathcal{D},\pmb{\beta})=\mathcal{T}\left(\Tilde{\mathbf{x}}\Big|\pmb{\mu}_N,\frac{\lambda_N+1}{\lambda_N(\nu_N-p+1)}\pmb{\Psi}_N,\nu_N-p+1\right),
\end{equation}
defined as
\begin{equation}
\label{eq:tDist}
    \mathcal{T}(\mathbf{x}|\pmb{\mu},\Sigma,\nu):=\frac{\Gamma(\nu/2+p/2)}{\Gamma(\nu/2)}\frac{|\Sigma|^{-1/2}}{(\pi\nu)^{p/2}}\left[1+\frac{1}{\nu}(\mathbf{x}-\pmb{\mu})^\top\Sigma^{-1}(\mathbf{x}-\pmb{\mu})\right]^{-\left(\frac{\nu+p}{2}\right)}.
\end{equation}

\subsection{Choice of prior hyperparameters}
From the properties of the NIW distribution~\eqref{eq:NIW}, there are conditions on the choice of prior hyperparameters. $\pmb{\Psi}$ must be a positive-definite real invertible $p\times p$ matrix and $\lambda>0$. We also must ensure that $\nu>p+1$ for expression~\eqref{eq:postPredTheta} to have a defined variance. These conditions will not be satisfied if one uses the (improper) uninformative prior, and in practice it is often better to use a weakly informative data-dependent prior, which avoids degeneracies and singularities. A common choice~\cite{chipman_practical_2001,fraley_bayesian_2007} is to use
\begin{equation}
\label{eq:priorQDA}
    \pmb{\Psi}=\frac{\mathrm{diag}(\mathbf{S})}{NK^{2/p}},\quad \pmb{\mu}_0=\Bar{\mathbf{x}},\quad\nu=p+2,\quad0<\lambda\ll 1. 
\end{equation}
This simplifies the posterior distribution~\eqref{eq:postQDA} with parameters
\begin{equation}
\label{eq:postHyp}
    \pmb{\Psi}_N=\frac{\mathrm{diag}(\mathbf{S})}{NK^{2/p}}+\mathbf{S},\,\pmb{\mu}_N=\Bar{\mathbf{x}},\,\nu_N=N+p+2,\,\lambda_N=N 
\end{equation}
\end{document}